\pdfoutput=1

\documentclass[11pt]{article}

\usepackage{acl}

\usepackage{times}
\usepackage{latexsym}

\usepackage[T1]{fontenc}

\usepackage[utf8]{inputenc}

\usepackage[most]{tcolorbox}
\usepackage{tikz}
\usetikzlibrary{shapes,calc,positioning}

\newcommand{\modelname}{\textsc{DocLens}\xspace}

\global
\tcbset{
  aibox/.style={
    width=\textwidth,
    top=10pt,
    colback=white,
    colframe=black,
    colbacktitle=black,
    enhanced,
    center,
    attach boxed title to top left={yshift=-0.1in,xshift=0.15in},
    boxed title style={boxrule=0pt,colframe=white,},
  }
}
\newtcolorbox{AIbox}[2][]{aibox,title=#2,#1}
\tcbset{
  aiboxsmall/.style={
    width=0.9\textwidth,
    top=10pt,
    colback=white,
    colframe=black,
    colbacktitle=black,
    enhanced,
    center,
    attach boxed title to top left={yshift=-0.1in,xshift=0.15in},
    boxed title style={boxrule=0pt,colframe=white,},
  }
}
\newtcolorbox{AIboxSmall}[2][]{aiboxsmall,title=#2,#1}
\definecolor{aigold}{RGB}{244,210, 1} 
\definecolor{aired}{RGB}{255,180,181}
\newlength\savewidth

\definecolor{defaultcolor}{gray}{0.9}

\usepackage{microtype}
\usepackage{comment}
\usepackage{todonotes}

\usepackage{amsmath,amsfonts,bm}
\usepackage{booktabs} %
\usepackage{url}
\usepackage{graphicx}
\usepackage{array}
\usepackage{color}
\usepackage{makecell}
\usepackage{multirow}
\usepackage{xspace}
\usepackage{colortbl}
\usepackage{subcaption}
\usepackage{algpseudocode}
\usepackage[noend,ruled]{algorithm2e}
\usepackage{setspace}
\usepackage{varwidth}
\usepackage{enumitem}
\usepackage{mdframed}
\usepackage{centernot}

\usepackage{soul}
\usepackage{pifont}
\usepackage{graphics}

\usepackage{cuted,tcolorbox,lipsum}

\newcommand{\start}[1]{\vspace{.3mm}\noindent{{\bf #1}.}}

\newcommand{\downv}{\vspace{-.0cm}}
\newcommand{\upv}{\vspace{-.0cm}}

\definecolor{gred}{RGB}{255,102,102}
\definecolor{gblue}{RGB}{51,102,255}
\definecolor{gyellow}{RGB}{244,180,0}
\definecolor{ggreen}{RGB}{15,157,88}
\definecolor{ggrey}{RGB}{115,115,115}
\definecolor{na}{gray}{0.9}

\definecolor{textRed}{RGB}{157,0,23}
\definecolor{textYellow}{RGB}{166,119,54}
\definecolor{textGreen}{RGB}{58,110,38}
\definecolor{textBlue}{RGB}{39,71,156}

\definecolor{LightYellow}{RGB}{255,250,208}
\definecolor{LightGreen}{RGB}{194,255,192}
\definecolor{LightBlue}{RGB}{187,236,251}
\definecolor{LightPurple}{RGB}{224,223,255}
\definecolor{LightGrey}{RGB}{225,225,225}

\definecolor{OrangeRed}{rgb}{1.0, 0.27, 0.0}
\definecolor{midnightgreen}{rgb}{0.0, 0.29, 0.33}
\definecolor{darkgreen}{rgb}{0.0, 0.42, 0.24}

\definecolor{diagramRed}{RGB}{246,193,193}
\definecolor{diagramPurple}{RGB}{224,224,253}
\definecolor{diagramOrange}{RGB}{244,222,176}

\newcommand{\colorR}[1]{\textcolor{gred}{\textbf{#1}}}
\newcommand{\colorG}[1]{\textcolor{ggreen}{\textbf{#1}}}

\usepackage{amsmath,amsfonts,bm}

\def\eqref#1{equation~\ref{#1}}

\def\1{\bm{1}}

\DeclareMathAlphabet{\mathsfit}{\encodingdefault}{\sfdefault}{m}{sl}
\SetMathAlphabet{\mathsfit}{bold}{\encodingdefault}{\sfdefault}{bx}{n}

\usepackage{cleveref}

\title{\modelname: Multi-aspect Fine-grained Evaluation for \\ Medical Text Generation}

\author{Yiqing Xie$^{1}$\thanks{~~Work done as an intern at Microsoft Research.} \quad Sheng Zhang$^{2}$ \quad Hao Cheng$^{2}$ \quad Pengfei Liu$^{3}$ \quad \bf Zelalem Gero$^{2}$ \\ 
\textbf{\quad Cliff Wong$^{2}$ \quad Tristan Naumann$^{2}$ \quad Hoifung Poon$^{2}$ \quad Carolyn Ros\'e$^{1}$}\\
$^{1}$Carnegie Mellon University 
\quad $^{2}$Microsoft Research 
\quad $^{3}$Shanghai Jiaotong University \\ 
}

\begin{document}
\maketitle
\begin{abstract}
Medical text generation aims to assist with administrative work and highlight salient information to support decision-making.
To reflect the specific requirements of medical text, in this paper, we propose a set of metrics to evaluate the completeness, conciseness, and attribution of the generated text at a fine-grained level.
The metrics can be computed by various types of evaluators including instruction-following (both proprietary and open-source) and supervised entailment models.
We demonstrate the effectiveness of the resulting framework, \modelname, with three evaluators on three tasks: clinical note generation, radiology report summarization, and patient question summarization. 
A comprehensive human study shows that \modelname exhibits substantially higher agreement with the judgments of medical experts than existing metrics.
The results also highlight the need to improve open-source evaluators and suggest potential directions.\footnote{We released the code at \url{https://github.com/yiqingxyq/DocLens}.}
\end{abstract}

\begin{figure*}[t]
\centering
    \centering
    \includegraphics[width=0.95\linewidth]{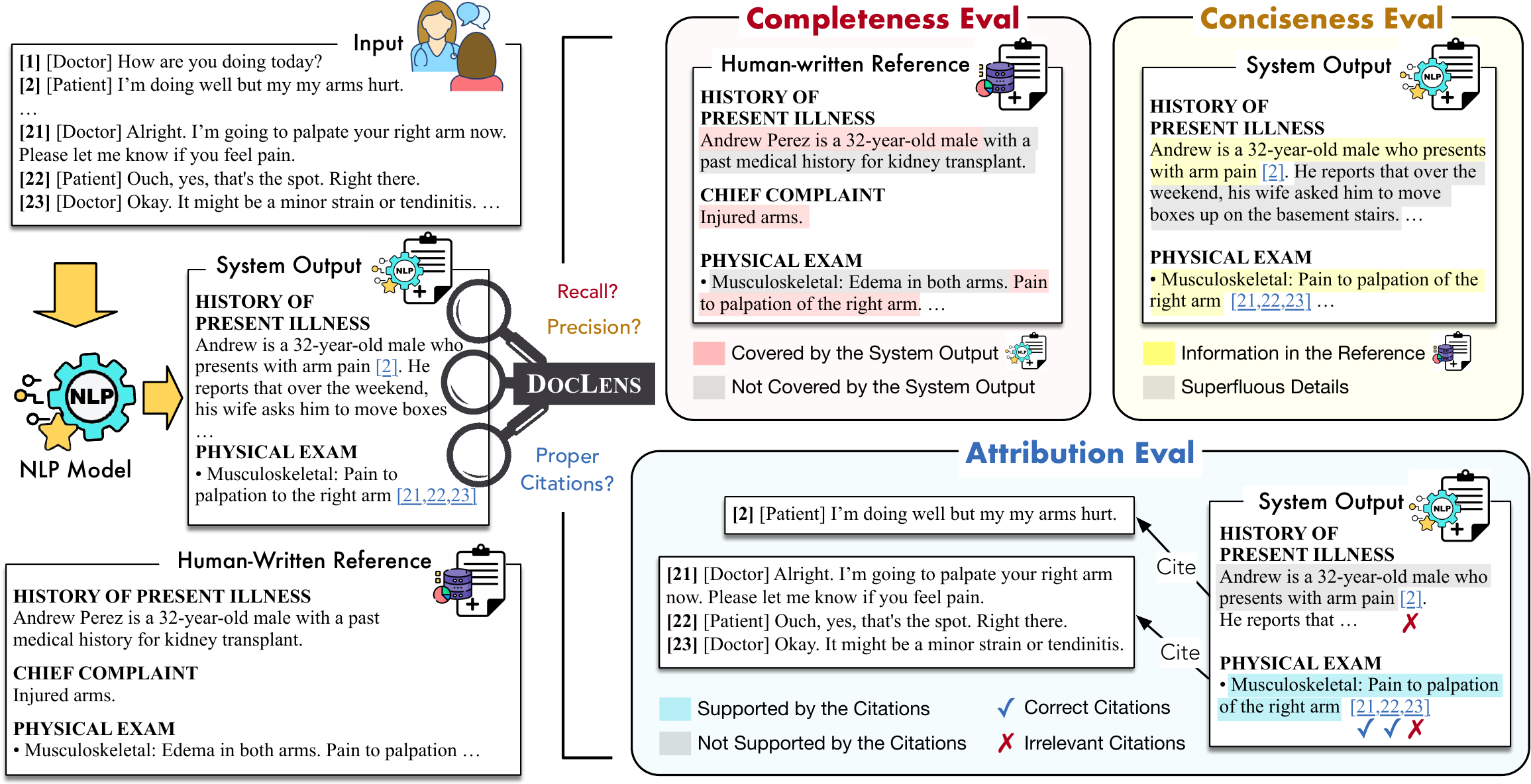}
    \upv
    \caption{Evaluation aspects of \modelname for medical text generation. \textcolor{textRed}{\textbf{Completeness}} evaluates the amount of salient details in the system output. \textcolor{textYellow}{\textbf{Conciseness}} evaluates the amount of information that is both accurate and salient. \textcolor{textBlue}{\textbf{Attribution}} checks whether the generated information can be traced back and attributed from the input}
    \label{fig:eval_aspects}
    \downv
\end{figure*}

\section{Introduction}

Medical text generation has been widely applied to various scenarios, including clinical note generation \cite{aci-bench,mts-dialog}, report summarization \cite{adams-etal-2021-whats,radadapt}, and patient question summarization \cite{MeQSum}. 
In report summarization, for example, text generation systems aim to assist medical experts by automatically summarizing the salient findings in a CT or MR report, which reduces the time on paperwork and supports decision-making~\cite{zhou-etal-2023-survey}.
To help medical experts decide whether to adopt text generation systems or which system to use, it is imperative to have a reliable evaluation methodology.

One line of work on medical evaluation conducts human evaluation under \emph{multiple aspects},
reflecting different criteria of an ideal generation result \cite{ben-abacha-etal-2023-empirical,zhou-etal-2023-survey,zhang-etal-2021-leveraging-pretrained}. 
To capture which exact information is inaccurate or omitted,
other human evaluation methods conduct more \emph{fine-grained} evaluations by examining each fact individually~\cite{ben-abacha-etal-2023-empirical}.
Due to the high cost and poor scalability of human evaluation, another line of work focuses on automatic evaluation. 
However, existing automatic medical evaluation methods typically assign a coarse-level score for the entire system output \cite{ben-abacha-etal-2023-investigation,zhou-etal-2023-survey}, without indicating the aspects or criteria the score reflects.
Recent general-domain evaluation methods focus on more fine-grained units such as sentences~\cite{ALCE} or atomic facts~\cite{factscore}.
However, existing methods neglect evaluation aspects critical to medical generation or require external knowledge sources in evaluation.

In this paper, we propose \modelname, which \emph{automatically} conducts evaluation of medical text generation at a fine-grained level, including both reference-based and reference-free aspects.
As shown in \autoref{fig:eval_aspects}, to evaluate the recall and precision of clinically significant information in the generation, we conduct a reference-based evaluation for the \textbf{completeness} and \textbf{conciseness} of the system-generated output.
Specifically, we break down the system output (e.g., generated clinical note) and reference (e.g., human-written note) into subclaims and assign a binary score for each subclaim.
In real-world scenarios, AI systems are typically used in an \emph{assisting} role, where the medical experts need to judge the reliability of the generated information as part of their process using the AI systems.
As a result, we evaluate \textbf{attribution}, which checks whether each piece of generated information is properly grounded in the input.
Specifically, we conduct a reference-free evaluation and check whether each generated sentence contains accurate references back to the input.

\modelname can be computed with various types of evaluator models and we experiment with three representatives: a proprietary model (GPT-4~\cite{openai2023gpt4}), an open-source instruction-following model (Mistral~\cite{mistral}), and a supervised natural language inference (NLI) model (TRUE~\cite{TRUE}). 
We apply \modelname with the three evaluators to benchmark multiple medical generation systems on three tasks: clinical note generation, radiology report summarization, and patient question summarization.
To compare the quality of \modelname with existing metrics and to compare different evaluators of \modelname,
we conduct a human study to investigate how well each metric aligns with medical experts' judgment.
Experiments show that \modelname exhibits substantially higher agreement with medical experts than existing metrics commonly used in the medical domain.
The results also reveal the gap between open-source and proprietary evaluators. Our analyses further suggest potential directions to improve open-source evaluators.

\start{Contributions}
(1) We identify crucial aspects of medical text evaluation and design corresponding metrics for conducting a fine-grained evaluation.
(2) We present an automatic evaluation framework, \modelname, based on the metrics, which can be computed by various types of evaluators.
(3) We apply \modelname to three medical generation tasks.
Human study results show that \modelname exhibits substantially higher agreement with human judgments than existing metrics.

\section{Related Work}
\start{Evaluation in the Medical Domain}
Existing approaches to medical text evaluation
~\cite{clinical-text-summ,radadapt,medpalmm} have adopted traditional metrics from general NLP, including n-gram-based metrics~\cite{rouge,bleu}, embedding-based methods~\cite{bertscore}, and model-based methods~\cite{bleurt}.
Other existing approaches evaluate the overlap of medical concepts \citet{factual-f1,image2text-report-summ,report-summ-semantic-rewards}, which utilizes information extraction models~\cite{radgraph} to extract clinical entities and relations from the system output and the reference and compute their overlap.
Such metrics heavily rely on surface-level similarities and lack validity.

\start{Factuality Evaluation in the General Domain}
Factuality evaluation is the most relevant topic to our work, as it judges the factual alignment between input and output~\cite{fever,liar,multifc}.
A common approach is to assign a single score for the entire system output \cite{liu2023geval,fu2023gptscore}.
This does not satisfy the needs of medical applications, where every piece of information is essential and requires careful examination.
Another line of work decomposes the system output into fine-grained units, such as content units~\cite{nenkova-passonneau-2004-evaluating}, short subclaims~\cite{wright-etal-2022-generating,chen-etal-2022-generating,wice} or atomic facts that each convey only one piece of information~\cite{factscore}.
Towards more transparent text generation, a strand of prior work further evaluates the attribution of generated facts by training or prompting the models to ground them with references back to the input~\cite{gao-etal-2023-rarr,ALCE,liu2023evaluating,AttrScore}.

To satisfy the specific requirements of the medical domain, our work extends past work by (1) evaluating the recall, precision, and attribution of the generated facts by conducting completeness, conciseness, and attribution evaluations at a fine-grained level, and (2) adapting a wider range of evaluator models to compute the metrics automatically and conducts empirical comparisons, providing a diverse range of selections.

\begin{figure}[!t]
\centering
    \centering
    \includegraphics[width=0.9\linewidth]{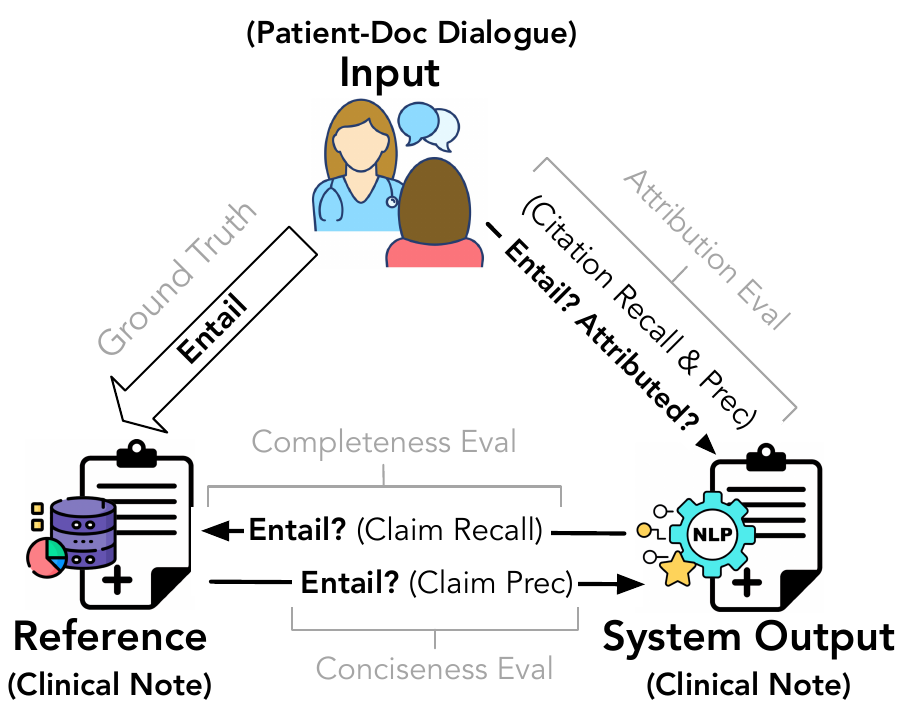}
    \upv
    \caption{To conduct a multi-aspect evaluation, we verify the entailment relations among the input (e.g., patient-doctor dialogue), system output (e.g., generated clinical note), and reference (e.g., human-written clinical note).
    }
    \label{fig:entail_diagram}
    \downv
\end{figure}

\begin{figure*}[t]
\centering
    \centering
    \includegraphics[width=0.95\linewidth]{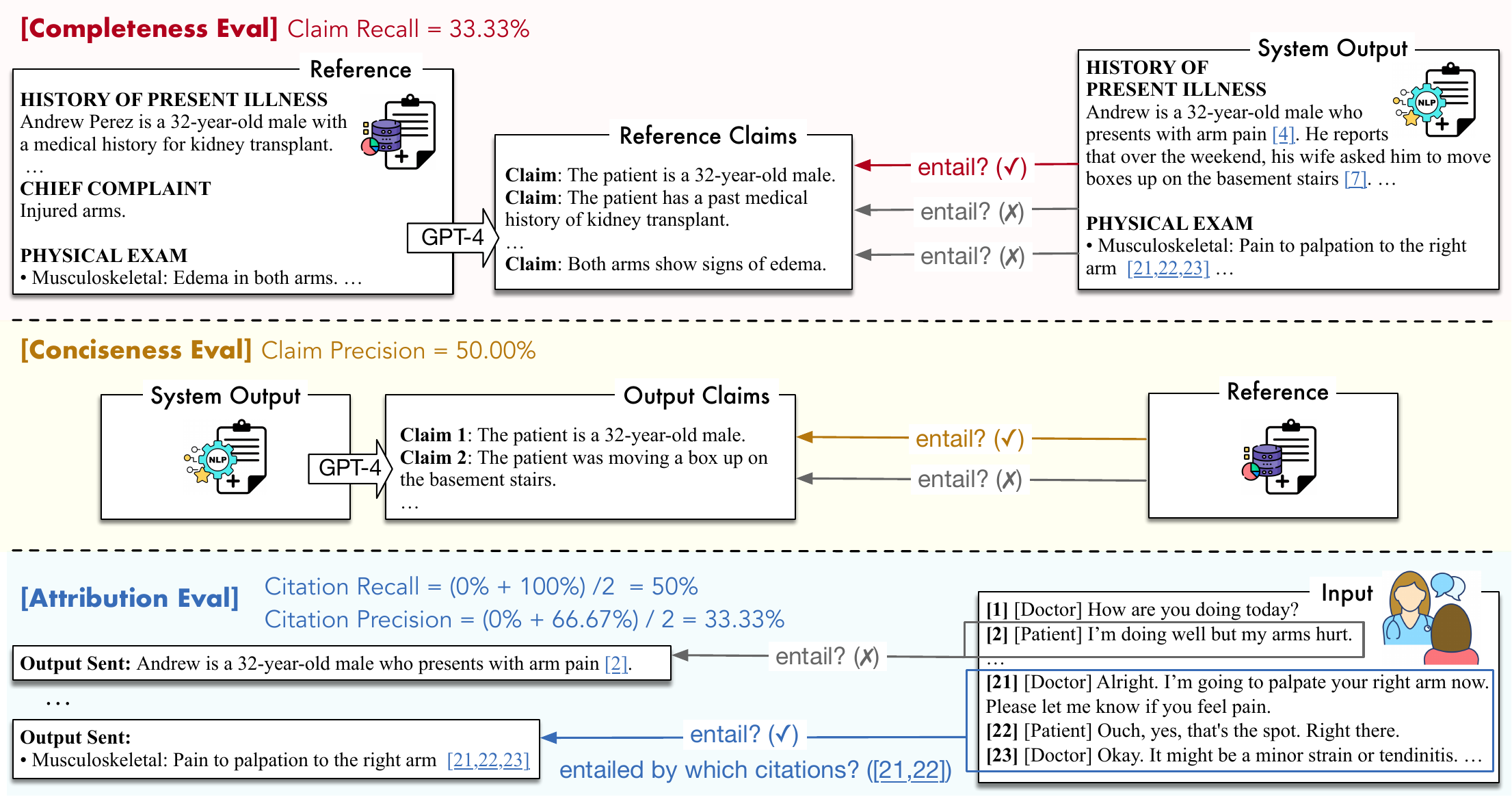}
    \upv
    \caption{Illustration of the metrics of \modelname for medical evaluation:
    \textcolor{textRed}{\textbf{Claim Recall}} measures the proportion of claims from the human-written reference that can be entailed by the system output. 
    \textcolor{textYellow}{\textbf{Claim Precision}} measures the proportion of claims from the output that can be entailed by the reference. 
    \textcolor{textBlue}{\textbf{Citation Recall}} measures the proportion of output statements that can be entailed by their corresponding citations. 
    \textcolor{textBlue}{\textbf{Citation Precision}} measures the proportion of citations that factually support the associated statement.}
    \label{fig:eval_pipeline}
    \downv
\end{figure*}

\section{Evaluation Framework of \modelname}
\label{sec:eval_framework}

There are two special characteristics of clinical text:
(a) \emph{Criticality}: every piece of information is essential. 
Medical documentation must be free from inaccuracies and omissions.
(b) \emph{Assistance}: the generated text will be examined by experts. As decisions can be life-critical, the generated text is positioned as a resource rather than a final product.

In line with these considerations, our proposed framework employs three evaluation aspects: completeness (\S\ref{sec:completeness}), conciseness (\S\ref{sec:conciseness}), and attribution (\S\ref{sec:attribution}) and design a set of corresponding metrics. 
The proposed metrics can be automatically computed using a variety of evaluator models (\S\ref{sec:evaluators}).
As shown in \autoref{fig:entail_diagram}, we formulate the three evaluation aspects as verification of entailment relations between the input, system output, and human-written references.
An illustrative example of each metric is shown in \autoref{fig:eval_pipeline}.

\subsection{Completeness Evaluation} 
\label{sec:completeness}
We first evaluate completeness: the amount of clinically significant information in the system output, which corresponds to the relation \textit{``System Output $\Rightarrow$ Reference''} in \autoref{fig:entail_diagram}. 
This can be viewed as the recall of the system output.
Based on the characteristic of \emph{criticality}, unlike previous work~\cite{clinical-text-summ} that assigns an overall score to the system output, we are also interested in which exact salient information is retained or omitted.

We introduce \textbf{claim recall} to evaluate completeness at a fine-grained level.
As shown in \autoref{fig:eval_pipeline}, we first break down the reference into a list of subclaims using GPT-4~\cite{openai2023gpt4}, where each subclaim states only one fact in the reference.
Let $y$ be the reference, $\mathcal{L}_y$ be the list of reference subclaims, $y'$ be the system output.
We apply an evaluator model to judge whether each claim $l \in \mathcal{L}_y$ is entailed by the generated output $y'$.
Claim recall is then formally defined as $\frac{1}{|\mathcal{L}_y|}\sum_{l \in \mathcal{L}_y} \mathbb{I}[ y' \Rightarrow l ]$.

The concept of claim recall parallels the definition provided in ALCE \cite{ALCE}, with the following difference:
ALCE only requires the output to follow the key reasoning steps in the reference and hence restricts the extraction to three claims per instance.
In contrast, since an ideal clinical note should cover all salient details, we prompt the model to generate claims that encapsulate all factual information in the reference.

\subsection{Conciseness Evaluation} 
\label{sec:conciseness}
Following the characteristic of \emph{assistance}, an ideal system output will allow medical experts to quickly capture salient information, without spending significant time on superfluous details.
We thus evaluate conciseness: the amount of generated information that is both factually accurate and salient.
Assuming that the reference only contains salient information, conciseness evaluation aligns with the \textit{``Reference $\Rightarrow$ System Output''} relation in \autoref{fig:entail_diagram}.

We define \textbf{claim precision} to evaluate conciseness. 
Similar to claim recall, we generate a list of claims for the system output and apply the evaluator to compute the proportion of claims that can be entailed by the reference.
Formally, let $y$ be the reference, $y'$ be the system output, and $\mathcal{L}_y'$ be the list of output subclaims, we define claim precision as $\frac{1}{|\mathcal{L}_{y'}|}\sum_{l \in \mathcal{L}_{y'}} \mathbb{I}[ y \Rightarrow l ]$. 

\subsection{Attribution Evaluation} 
\label{sec:attribution}
Based on the characteristic of \emph{assistance}, to help medical experts quickly verify the output statements, we also evaluate attribution: the amount of generated information that can be traced back from the input, which corresponds to the \textit{``Input $\Rightarrow$ System Output''} relation in \autoref{fig:entail_diagram}.
As shown in \autoref{fig:eval_pipeline}, the system also generates citations to the input following each statement, which helps medical experts quickly locate the relevant context and verify the generated information.

Following existing work \cite{liu2023evaluating,ALCE}, we first compute \textbf{citation recall}, which evaluates whether each statement in the system output can be fully supported by the combination of its cited sentences.
Formally, for each output statement $s$, let $\mathcal{C}$ be the set of input sentences it cites. The citation recall of $s$ is $1$ if and only if $\mathcal{C} \Rightarrow s$ and otherwise $0$.
The citation recall of the whole output is then defined as the percentage of statements that can be entailed by their citations.

We also evaluate \textbf{citation precision} to examine whether the system generates redundant citations.
Intuitively, if a statement $s$ can be supported by the combination of its citations $\mathcal{C}$, a citation $c \in \mathcal{C}$ is necessary if it independently supports the statement $s$, or if removing it leaves the statement unsupported.
Formally, we define the citation precision of $c$ as $1$ if and only if:

$(i)$ $\mathcal{C} \Rightarrow s$, and

$(ii)$ $c \Rightarrow s$ or $\mathcal{C}\setminus\{c\} \centernot \Rightarrow s$.

\noindent For instance, in \autoref{fig:eval_pipeline}, the output statement cites conversational turns ``\texttt{[21][22][23]}'' in the input, but only \texttt{[21]} and \texttt{[22]} are pertinent to the output. So \texttt{[21]} and \texttt{[22]} will have citation precision $= 1$ and \texttt{[23]} will have citation precision $= 0$.
We define the citation precision of the whole output as the average precision of all its citations.

\subsection{Discussion of Excluded Aspects} 
\label{sec:other_aspects}
There are other evaluation aspects for text generation in general domains, such as \textbf{coherence} and \textbf{fluency}~\cite{zhong-etal-2022-towards,ALCE}, where coherence evaluates whether the generated sentences form a coherent body and fluency evaluates whether each sentence is well-written and grammatical.
These aspects are less emphasized in the medical domain since they do not directly impact treatment outcomes.

To ensure the accuracy of the generated text, many existing methods evaluate \textbf{factual consistency}~\cite{fever,multifc,factscore}, which compares the factual statements in the generated text and the input.
Among our proposed aspects, conciseness and attribution both incorporate the need for factual consistency.
In addition to the factuality of output statements, conciseness further evaluates whether the statements are salient and attribution judges whether they can be traced back from the input.

\subsection{\modelname with Various Evaluators}
In this section, we introduce how we compute the metrics of \modelname with various evaluator models, including NLI models and open-source and proprietary instruction-following models.

\label{sec:evaluators}
\start{\modelname computed with NLI models}
Let $\phi(p,h)$ be the output of the NLI model, which is 1 if the premise $p$ entails the hypothesis $h$ and 0 otherwise. The claim recall, claim precision, and citation recall can be computed by $\phi($system output, reference claim$)$, $\phi($reference, output claim$)$, and $\phi($combination of citations, output statement$)$.

To compute citation precision for each citation $c\in\mathcal{C}$ in statement $s$, following our definition, $c$ has citation precision $= 1$ if and only if s has citation recall $= 1$ and $\phi(c,s) || \phi(\mathcal{C}\setminus\{c\}, s) = 1$.

\start{\modelname computed with instruction-following models}
We also apply instruction-following models to compute the metrics, including proprietary and open-source models.
To compute claim recall and claim precision, we prompt the evaluator to generate ``1'' or ``0'' for each claim based on whether they can be supported.

As for citation recall and precision, to reduce computation, we prompt the evaluator to predict if a statement is entailed by its citations and to identify the supporting citations in a single call.
In the example of \autoref{fig:eval_pipeline}, where the citations ``\texttt{[21][22][23]}'' support the output statement but \texttt{[23]} is irrelevant, the evaluator should output ``1'' for the entailment prediction and ``\texttt{[21][22]}'' as the supporting citations.

We further adopt two prompt styles to improve the quality of instruction-following evaluators, where the example prompts are shown in \S\ref{sec:evaluator_prompts}:

(1) \textbf{Generation with Structure}.
Existing research has observed that models have better performance when they are prompted to generate in a structured format, such as logic representations, or pseudo code \cite{mishra2023pseudocode}.
With the same high-level idea, we prompt the evaluator to generate the entailment prediction in a JSON dictionary.

(2) \textbf{Chain-of-Thought}.
Chain-of-thought (CoT) prompts the model to generate a series of intermediate reasoning steps, which has shown to be effective in various tasks~\cite{cot}.
Following this idea, we prompt the evaluator to generate the explanation before the prediction. 

We conduct experiments on two NLI datasets: MedNLI~\cite{MedNLI} and ANLI~\cite{ANLI} to investigate the effectiveness of the two prompt styles on predicting entailment relationships. MedNLI evaluates reasoning with medical knowledge and ANLI focuses on commonsense reasoning. Both are reasoning abilities that an evaluator needs.

To align with our evaluation setting, we adopt a 2-way classification setting where ``entailment'' forms one class and both ``neutral'' and ``contradiction'' are merged into the other class.
As shown in \autoref{tab:NLI_new}, the GPT-4 evaluator benefits the most from the combination of generation in JSON and CoT, and the Mistral evaluator only benefits from CoT. We observe that in many cases, Mistral fails to generate outputs in a valid JSON format, which leads to parsing error when reading the results.
Detailed experiment results are shown in \S\ref{sec:prompt_style_experiments}.

\begin{table}[t]
\centering
\resizebox{\linewidth}{!}{
\begin{tabular}{lccc}
\toprule
\bf Model & \bf MedNLI & \bf ANLI & \bf Weight AVG \\
\midrule
TRUE & 81.9 & 71.5 & 74.3 \\
\midrule
GPT-4 (2-shot) & \bf \underline{92.8} & 86.1 & 87.8 \\
\quad + JSON & 91.0 & \bf \underline{87.7} & 88.5 \\
\quad + CoT & 91.6 & 86.7 & 88.2 \\
\quad + JSON + CoT & 91.8 & 87.5 & \bf \underline{88.6} \\
\midrule
Mistral (2-shot) & 84.8 & 69.6 & 73.4 \\
\quad + JSON & \underline{87.8} & 67.8 & 72.8 \\
\quad + CoT & 87.2 & \underline{70.6} & \underline{74.8} \\
\quad + JSON + CoT & 87.3 & 70.0 & 74.3 \\
\bottomrule
\end{tabular}
}
\caption{Classification accuracy on MedNLI and ANLI under the 2-way classification setting.
``ANLI'' is the average accuracy on ANLI (R1, R2, R3).
We provide one in-context example from each of the 2 classes.
}
\label{tab:NLI_new}
\end{table}

\begin{figure*}[ht]
  \centering
  \includegraphics[width=\linewidth]{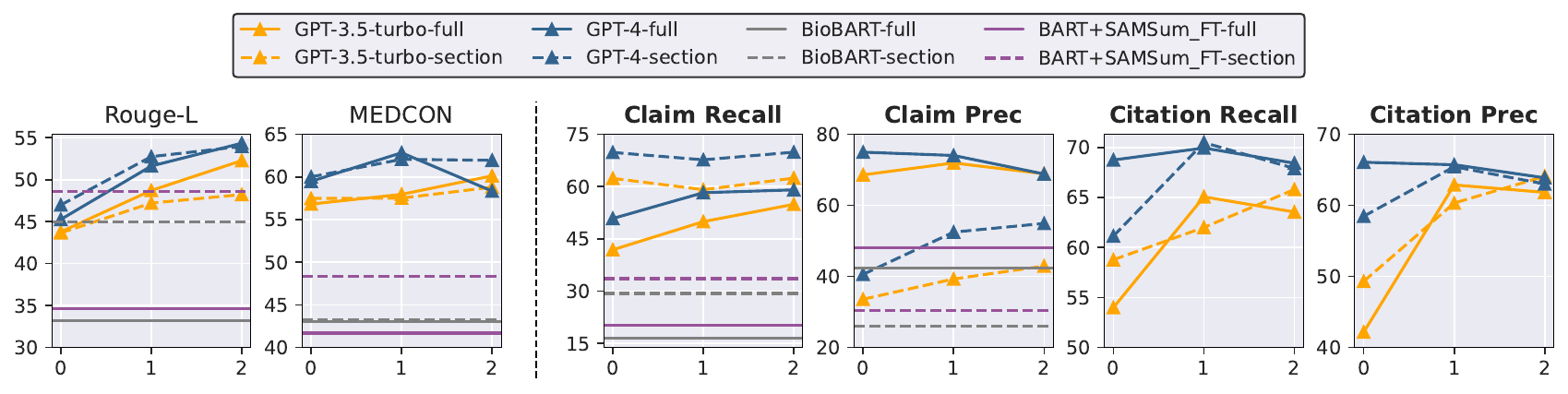}
  \caption{Clinical note generation results on ACI-BENCH~\cite{aci-bench}.
  We split the results under existing metrics and \modelname computed with GPT-4.
 We evaluate open-source and proprietary note generation models with different numbers of in-context examples.
 }
  \label{fig:aci_performance}

  \includegraphics[width=\linewidth]{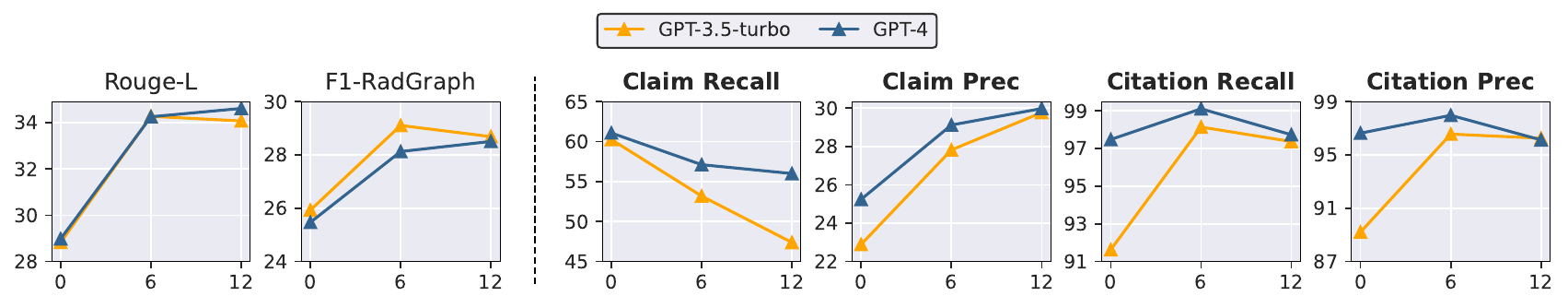}
  \caption{Radiology report summarization results on MIMIC-III~\cite{radadapt} evaluated by existing metrics and \modelname computed with GPT-4.}
  \label{fig:mimic_performance}

  \includegraphics[width=\linewidth]{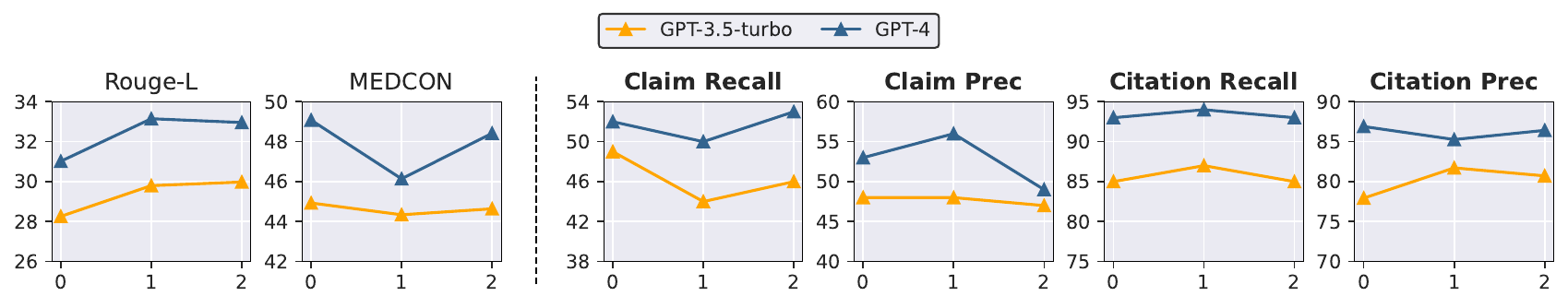}
  \caption{Patient question summarization results on MeQSum~\cite{MeQSum} evaluated by existing metrics and \modelname computed with GPT-4.}
  \label{fig:meqsum_performance}
\end{figure*}

\begin{figure*}[t]
    \centering
    \includegraphics[width=\linewidth]{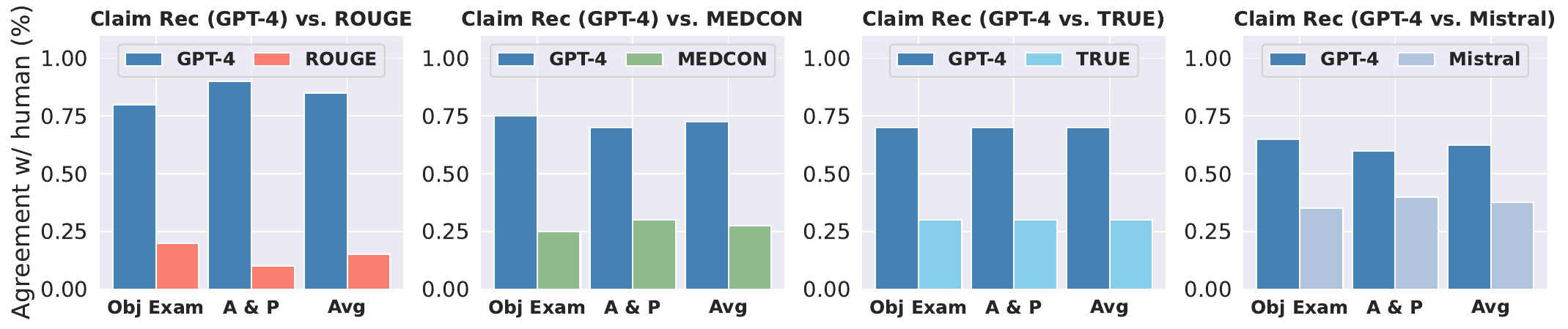}
\upv
\caption{Agreement between each metric and the \emph{subjective preferences} of medical experts over two system outputs.
We only annotate the system outputs pairs \emph{where the two metrics have different preferences}.
The outputs are selected from the Objective Exam (O-Exam) and Assessment and Plan (A\&P) section in note generation.
}
\downv
\label{fig:subj_preference}
\end{figure*}

\section{Experiments}
\label{sec:exp}
In this section, we aim to answer three research questions: 
(\textbf{RQ1}) How do different medical generation models perform under our evaluation?
(\textbf{RQ2}) How is the evaluation quality of \modelname compared to existing metrics?
(\textbf{RQ3}) How is the evaluation quality of \modelname computed with open-source evaluators compared to proprietary ones?

\subsection{Evaluation tasks and metrics}
\label{sec:tasks}
To answer \textbf{RQ1}, we experiment with three types of evaluators for \modelname: proprietary and open-source instruction-following models, and NLI models, and evaluate three representative medical generation tasks: clinical note generation~\cite{aci-bench}, medical report summarization~\cite{radadapt}, and patient question summarization~\cite{MeQSum}.

\start{Clinical note generation}
Clinical note generation is defined as generating a ``SOAP'' note given a dialogue between a doctor and a patient~\cite{aci-bench,mts-dialog}. 
A SOAP note consists of the subjective, objective Exam, objective results, and assessment and plan sections. 
We conduct experiments on the ACI-BENCH dataset~\cite{aci-bench}. A test example is shown in \autoref{tab:aci_example}.

\start{Radiology report summarization}
We follow the setting from~\newcite{radadapt}, where the input is the findings section of a radiology report, and the goal is to generate an impression section that contains key observations and conclusions. 
we utilize the public test set of MIMIC-III~\cite{mimic}.
A test example is shown in \autoref{tab:mimic_example}.

\start{Patient question summarization}
Question summarization aims to generate a condensed question expressing the minimum information required to find correct answers to the original question~\cite{MeQSum}.
We utilize the MeQSum dataset, which consists of consumer health questions and their corresponding summaries authored by medical experts.
A test example is shown in \autoref{tab:meqsum_example}.

\start{Evaluation methods} 
We evaluate the three tasks with \modelname computed with GPT-4~\cite{openai2023gpt4}, Mistral~\cite{mistral}, and TRUE~\cite{TRUE}, representing three types of models.
We also apply existing commonly-used metrics~\cite{aci-bench,radgraph,clinical-text-summ}. 
We list the metrics for each task in \autoref{tab:metric_list}.

Experimental details are provided in \S\ref{sec:exp_details}.

\subsection{Exp-I: Medical Text Generation Results}
\label{sec:results}

\Cref{fig:aci_performance,fig:mimic_performance,fig:meqsum_performance} show the results on the three representative tasks evaluated by existing metrics and \modelname computed with GPT-4.
The detailed numbers are provided in \Cref{tab:ACI_existing,tab:ACI_combine,tab:MIMIC_existing,tab:MIMIC_orig,tab:meqsum_existing,tab:meqsum}. 
We show the results of \modelname with the Mistral evaluator in \Cref{tab:ACI_Mistral,tab:MIMIC_Mistral,tab:meqsum_Mistral} and the TRUE evaluator in \Cref{tab:ACI_TRUE,tab:MIMIC_TRUE,tab:meqsum_TRUE}.

\start{Influence of in-context examples}
We observe that the results under most metrics can be improved by adding in-context examples, but increasing the number of examples from 1 to 2 leads to diminishing returns.
Intriguingly, when the prompt only contains the instruction with no examples, GPT-3.5-turbo often fails to generate any citations or produces all citations together at the end. 
In contrast, GPT-4 consistently generates citations in the correct format across the three datasets.

\start{Proprietary vs. Open-source generation models} As shown in \autoref{fig:aci_performance}, GPT-based models outperform open-source models in the majority of experiments. We also observe that open-source models typically generate much shorter outputs than GPT-based models with heavy omission. E.g., BART+SAMSum (full) generates 179.4 words on average and GPT-4 (full, 2-shot) generates 351.9.

\start{Results under different evaluators}
Comparing results computed by \modelname with different evaluators,
we can observe that Mistral assigns overall higher scores than the other models, which in many cases misjudges ``partially support'' as ``fully support''.
We can also observe that the correlation between TRUE and GPT-4 is much lower on MeQSum than the other two datasets. The reason might be TRUE is mainly trained on declarative sentences and hence has unsatisfactory performance in judging the entailment relationships between questions.

\begin{table}[t]
\centering
\resizebox{\linewidth}{!}{
\begin{tabular}{lcccc}
\toprule
\multirow{3}{*}{\bf Comparison} 
& \multicolumn{4}{c}{\bf Corr w/ Medical Experts} \\
\cmidrule(lr){2-5}
& \multicolumn{2}{c}{\bf O-Exam} 
& \multicolumn{2}{c}{\bf A \& P} \\
\cmidrule(lr){2-3} \cmidrule(lr){4-5}
& $\rho$ & $\tau$ & $\rho$ & $\tau$ \\
\midrule
\midrule
Rouge-L Recall & 0.326 & 0.267 & -0.389 & -0.307 \\
Claim Recall (GPT-4) & \bf 0.787 & \bf 0.653 & \bf 0.732 & \bf 0.638 \\
\midrule
\midrule
MEDCON Recall & 0.138 & 0.103 & 0.132 & 0.078 \\
Claim Recall (GPT-4) & \bf 0.752 & \bf 0.621 & \bf 0.820 & \bf 0.652 \\
\midrule
\midrule
Claim Recall (TRUE) & 0.710 & 0.526 & 0.251 & 0.168 \\
Claim Recall (GPT-4) & \bf 0.953 & \bf 0.844 & \bf 0.522 & \bf 0.431 \\
\midrule
\midrule
Claim Recall (Mistral) & 0.627 & 0.486 & 0.342 & 0.234 \\
Claim Recall (GPT-4) & \bf 0.682 & \bf 0.612 & \bf 0.702 & \bf 0.546 \\
\bottomrule
\end{tabular}
}
\caption{
Spearman ($\rho$) and Kendall-$\tau$ correlation between each recall-based metric and the completeness scores assigned by medical experts. 
When comparing two metrics, we only annotate the system outputs pairs \emph{where the two metrics have different preferences}.
}
\label{tab:human_spearman}
\end{table}

\begin{figure}[t]
  \centering

  \begin{subfigure}[b]{0.245\textwidth}
    \includegraphics[width=\textwidth]{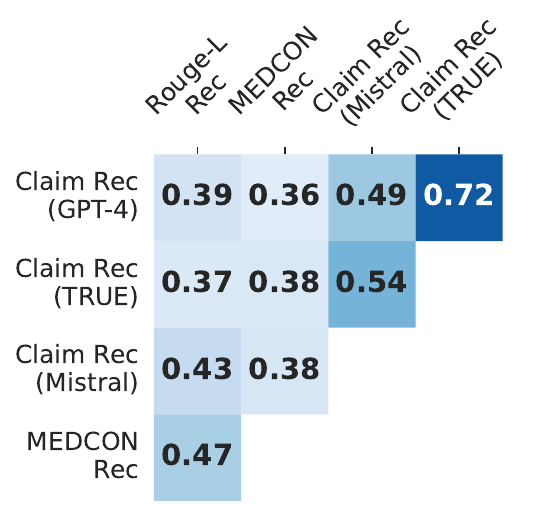}
    \caption{Recall-based metrics}
  \end{subfigure}
  \hspace{-0.4cm}
  \begin{subfigure}[b]{0.245\textwidth}
    \includegraphics[width=\textwidth]{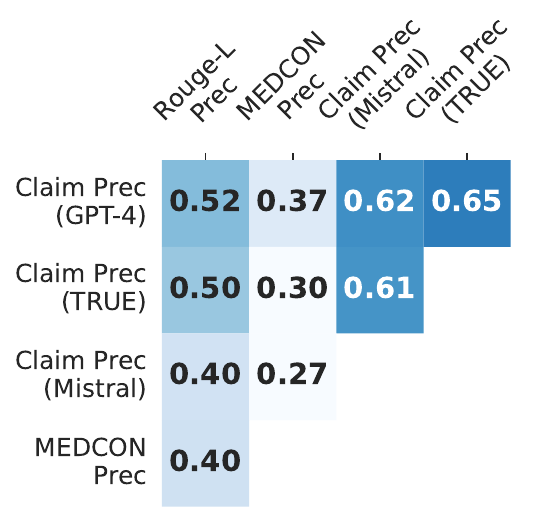}
    \caption{Precision-based metrics}
  \end{subfigure}
  
  \caption{Kendall's $\tau$ coefficients between recall-based and precision-based metrics on note generation.
  }
  \label{fig:metric_corr}
\end{figure}

\subsection{Exp-II: Agreement with Human}
\label{sec:human-preference}
To answer (\textbf{RQ2}) and (\textbf{RQ3}), we conduct a human study to check the alignment of different evaluation methods with medical experts.

\start{Setup} 
We focus on the completeness evaluation and compare claim recall computed by GPT-4 with other metrics, including (i) existing recall-based metrics: Rouge-L recall and MEDCON recall, and (ii) claim recall computed by Mistral and TRUE.

As shown in \autoref{tab:disagreement_pairs}, we observe that the metrics have agreed preferences in most of the cases. As our primary goal is to compare pairs of metrics, to reduce the required amount of human annotation, we only annotate pairs of system outputs \textit{where the two metrics disagree}: one metric ranks one system output higher, and the other metric ranks the other output higher.
We select system outputs from the ``Objective Exam'' (O-Exam) and ``Assessment and Plan'' (A\&P) sections in clinical note generation.

We invite five medical experts to the human study and assign three of them to label each pair of selected system outputs, without telling them which model generates which output
Given the reference note and the reference subclaims, we ask the medical experts (1) what percent of the claims can be entailed by each output, and (2) which output they think is more complete (i.e., their \emph{subjective preference}).
We present the inter-annotator agreement when comparing each two metrics in \autoref{tab:inter_annotator}. The average Spearman ($\rho$) coefficient is 0.763. More details are shown in \S\ref{sec:annotation_details}.

\start{\modelname vs. existing metrics}
\autoref{tab:human_spearman} shows the Spearman ($\rho$) and Kendal-$\tau$ correlations between metrics and medical experts and \autoref{fig:subj_preference} shows the agreement with human subjective preference.
Results show that in both experiments, claim recall (GPT-4) has a substantially better alignment with human than Rouge-L and MEDCON, which typically misjudges the cases where the system output and the reference have little or no lexical overlap.
An example is shown in \autoref{tab:case_rouge}.

\start{Comparison among \modelname evaluators}
We can also observe from \autoref{tab:human_spearman} and \autoref{fig:subj_preference} that there is still a large gap between open-source models (Mistral and TRUE) and GPT-4. 
As shown in \autoref{tab:case_Mistral}, we observe several patterns in the cases where GPT-4 and Mistral disagree:
\textbf{[Case 1]} The judgment requires medical knowledge. In claim 1, a medical expert would know that ``\texttt{lungs are clear}'' already means there are ``\texttt{no wheezes, rales, or rhonch}'' and hence ``\texttt{clear bilaterally}'' fully support this claim. However, this is misjudged by Mistral, which suggests that continuous pretraining on medical corpus could be beneficial.
\textbf{[Case 2]} The output only partially entails the subclaim.
Although output 2 mentions the ``\texttt{systolic ejection murmur}'', it omits the fact that the murmur is unchanged, but Mistral does not capture the omission.
\textbf{[Case 3]} Multiple facts are related to each other. Output 2 omits the fact of ``\texttt{no masses}'' in claim 3. However, Mistral wrongly predicts claim 3 as ``supported'' with the explanation: The notes states that there is no tenderness, which could potentially indicate that there is no masses.
In both case 2 and 3, Mistral does not strictly follow the instruction of ``judge whether the text fully supports the claim'', which calls for instruction tuning for entailment.

\begin{table}[!t]
\centering

\resizebox{\columnwidth}{!}{
\begin{tabular}{p{10.1cm}}
\toprule
\colorbox{LightGrey}{\textit{\textbf{Subclaims of the reference} (Used to compute claim recall)} \hskip2.7em } \\
1. Lungs are clear bilaterally, with no wheezes, rales, or rhonchi. \\ 
2. The patient has a grade 2/6 systolic ejection murmur, unchanged. \\ 
3. Examination of the abdomen shows no masses or tenderness. \\ 

\midrule 

\colorbox{LightGrey}{\textit{\textbf{Output 1} (Preferred by GPT-4 and human)} \hskip9.3em } \\
PHYSICAL EXAMINATION \\ 
The doctor performs a physical exam and finds the patient's lungs to be clear bilaterally and no tenderness or pain in the abdomen. The patient has a grade two out of six systolic ejection murmur in her heart, which has not changed since the previous visit.
\\
\\
\textbf{Claim Recall (Mistral)}: \colorR{33.34} \quad // Support Claim 2.  \\
\textbf{Claim Recall (GPT-4)}: \colorG{66.67} \quad // Support Claim 1 and 2. \\
\textbf{Human Judgment}: \colorG{66.67} \quad // Support Claim 1 and 2. \\
\midrule

\colorbox{LightGrey}{\textit{\textbf{Output 2} (Preferred by Mistral)} \hskip13.7em} \\
PHYSICAL EXAMINATION \\ 
• Lungs: Clear bilaterally. No wheezes, rales, or rhonchi. \\ 
• Heart: Grade 2/6 systolic ejection murmur. \\ 
• Abdomen: No tenderness to palpation. \\ 
\\
\textbf{Claim Recall (Mistral)}: \colorG{100.00} \quad // Support Claim 1, 2 and 3. \\
\textbf{Claim Recall (GPT-4)}: \colorR{33.34} \quad // Support Claim 1. \\
\textbf{Human Judgment}: \colorR{33.34} \quad // Support Claim 1. \\
\midrule
\textbf{Preference of Human \& claim recall (GPT-4)}: \textit{Output 1} \\
\textbf{Preference of claim recall (Mistral)}: \textit{Output 2} \\
\bottomrule
\end{tabular}
}
\caption{An example of disagreement between claim recall (GPT-4) and claim recall (Mistral) on the completeness over two system outputs.
}
\label{tab:case_Mistral}
\end{table}

\subsection{Exp-III: Correlations between Metrics}
\S\ref{sec:human-preference} shows that \modelname (GPT-4) aligns better with human than existing metrics or open-source evaluators.
To further study (\textbf{RQ2}) and (\textbf{RQ3}), we investigate to which extent these metrics diverge from each other.
Specifically, we divide the metrics into two groups: recall-based metrics (including claim recall, ROUGE-L recall, etc.) and precision-based metrics (including claim precision, ROUGE-L precision, etc.). We then compute Kendall's $\tau$ coefficient between each two metrics in each group.

Results in \autoref{fig:metric_corr} suggest that \modelname computed with various models exhibit relatively weak correlations with existing metrics, where the $\tau$ coefficients are lower than 0.5 in the majority of cases. Recall that \modelname (GPT-4) also has a better alignment with human when it disagrees with existing metrics, the results suggest that \modelname can highly improve current evaluation qualities.

We can also observe that although computing the same metrics, the correlations between \modelname computed with open-source evaluators and GPT-4 are not particularly high. 
For instance, there are only around 78\% of the system output pairs where Mistral and GPT-4 have the same preference.
This suggests that open-source evaluators exhibit significant divergence from proprietary models, which calls for future improvement.

\section{Conclusions and Future Works}
We present \modelname, a medical evaluation framework that judges three aspects with a set of fine-grained level metrics.
\modelname can be computed with various types of evaluator models, including proprietary and open-source instruction-following models and NLI models.
We apply \modelname to three tasks: clinical note generation, radiology report summarization, and patient question generation.
Human study shows that \modelname exhibits substantially higher agreement with human judgments than existing metrics. The results also reveals the substantial gap between proprietary and open-source evaluators. 

To close the gap, as suggested by our case study, future works could improve open-source evaluators by
(1) continuous pretraining the model on medical corpora, and
(2) instruction-tuning the model for entailment, where we can construct training data by adapting existing NLI datasets or leveraging the model itself to generate silver labels.
Another potential direction is to train the evaluator to generate multiple forms of feedback, such as the explanation of its judgement used in this work. Then evaluator model can then be applied to further improve medical text generation models.

\section{Acknowledgement}
We thank all participants in our human study for their hard work.
This work was supported in part by NSF grant DSES  2222762.

\section{Limitations}
We have only tested \modelname on public datasets, and hence this work cannot be directly used in real-world clinical scenarios.

\start{Self-bias of GPT-4}
While recent work~\cite{llm-as-a-judge} shows that there is no evidence that LLMs exhibit a self-enhancement bias, we agree that the GPT-4 evaluator may potentially have self-bias towards the text that is also generated by GPT-4. 
We use two ways in our paper to mitigate the potential bias; (1) implement two other evaluators: Mistral and TRUE, and (2) conduct a human study to verify the quality of the evaluators. As shown by our human study, GPT-4 still has a higher correlation with human judgment than the other two evaluators.

\start{Improvement of open-source evaluators}
In this work, we use the open-source evaluator models in a zero-shot or few-shot way, and observe that there is still a substantial gap between open-source evaluators and GPT-4. 
Given that GPT-4 is costly, inefficient, and potentially contains self-bias, future works could focus on further training open-source evaluators to bridge the gap.

\start{Medical question-answering}
Though we have evaluated our work on multiple medical summarization tasks, we have not conducted experiments on medical question-answering (QA), which is another crucial task in the medical domain.
The major reason is that current medical QA datasets mainly focus on short answers or multiple-choice questions~\cite{jin2019pubmedqa,medqa,medmcqa}, where evaluation is simpler than with the tasks we have focused on because in this case, it is possible simply to evaluate the exact match between the generated output and the answer.

\start{Multimodal medical generation}
Another limitation is that we do not consider the evaluation of multimodal medical generation, including visual QA~\cite{medvqa-survey,he2020pathvqa} and multimodal note/report generation. An interesting extension for future work may be to focus on the consistency between input and output in different modalities.

\section{Ethics Statement}
\start{License}
In our use of three public datasets, we have observed the highest ethical standards of conduct.  The specific datasets include: ACI-BENCH~\cite{aci-bench}, MIMIC-III~\cite{mimic}, and MeQSum~\cite{MeQSum}. The ACI-BENCH data is published under a Creative Commons Attribution 4.0 International Licence (CC BY). MIMIC-III is under the PhysioNet Credentialed Health Data License 1.5.0. MeQSum is distributed under the apache-2.0 license.

\start{Potential Risks}
Our framework leverages GPT-4 to evaluate medical data, which could be highly sensitive. To prevent data leakage as we have done, the potential users of our framework may use Azure OpenAI services in a HIPAA-compliant manner, which sets the privacy rule, the security rule, and the breach notification rule to protect patient health information\footnote{Details can be found in \url{https://learn.microsoft.com/en-us/azure/compliance/offerings/offering-hipaa-us}.}. The privacy rule especially imposes restrictions on the use and disclosure of patient health information without patient authorization.

\bibliography{anthology,custom}

\clearpage
\appendix

\section{Appendix}
\label{sec:appendix}

\subsection{Experiments on LLM Evaluators with Different Prompt Styles}
\label{sec:prompt_style_experiments}
In this section, we examine how different prompt styles affect the entailment ability of instruction-following evaluators.

\start{Experiments on existing NLI datasets}
We first conduct experiments on two prevalent natural language inference (NLI) datasets: ANLI~\cite{ANLI} and MedNLI~\cite{MedNLI}.
We use GPT-4 as the base model.
To evaluate binary factual consistency, we consider ``\emph{entailment}'' as one class and merge two NLI labels ``\emph{neutral}'' and ``\emph{contradiction}'' into the other class, denoting inconsistency between a premise-hypothesis pair.

Table~\ref{tab:NLI} compares supervised NLI and instruction-following models on ANLI and MedNLI under the 2-way classification setting.
GPT-4 significantly outperforms open-source models on both datasets and both generating in JSON format and CoT improves its performance.

We further compare GPT-4 and Mistral with supervised state-of-the-art models on MedNLI in Table~\ref{tab:NLI_3way}. 
With few-shot examples, GPT-4 outperforms supervised models, but Mistral does not have satisfactory performance. 

\begin{table*}[t]
\centering
\resizebox{0.8\linewidth}{!}{
\begin{tabular}{lccccc}
\toprule
\bf Model / Prompt Style $\downarrow$ 
& \bf ANLI-R1 & \bf ANLI-R2 & \bf ANLI-R3 & \bf MedNLI 
& \bf Average \\ 
\bf |Test Set| & 1000 & 1000 & 1200 & 1422 & -- \\
\midrule
TRUE & 78.8 & 69.1 & 67.3 & 81.9 & 74.3 \\
\midrule
Mistral (2-shot) & 68.9 & 69.2 & 70.8 & 84.8 & 73.4 \\
\quad + JSON & 68.2 & 65.6 & 69.7 & 87.8 & 72.8 \\
\quad + CoT & 70.6 & 69.6 & 71.7 & 87.2 & 74.8 \\
\quad + JSON + CoT & 71.5 & 66.8 & 71.7 & 87.3 & 74.3 \\
\midrule
GPT-4 (2-shot) & 88.7 & 84.5 & 85.1 & \bf 92.8 & 87.8 \\
\quad + JSON & \bf 91.4 & \bf 86.3 & 85.4 & 91.0 & 88.5 \\
\quad + CoT & 90.0 & 84.9 & 85.2 & 91.6 & 88.2 \\
\quad + JSON + CoT & 90.6 & 86.2 & \bf 85.6 & 91.8 & \bf 88.6 \\
\bottomrule
\end{tabular}
}
\caption{Accuracy on ANLI and MedNLI under the 2-way classification setting. We combine ``neutral'' and ``contradiction'' into one class. In the 2-shot prompt, we provide one example for each class.}
\label{tab:NLI}
\end{table*}

\begin{table}[t]
\centering
\resizebox{0.9\linewidth}{!}{
\begin{tabular}{lc}
\toprule
\bf Model / Prompt Style $\downarrow$ 
& \bf MedNLI \\
\midrule
Supervised & \\
\quad T5-large \cite{phan2021scifive} & 83.8 \\
\quad ClinicalT5-large \cite{clinicalt5} & 85.9 \\
\quad BioBART-large \cite{yuan-etal-2022-biobart} & 86.3 \\
\quad SciFive-large \cite{phan2021scifive} & 86.6 \\
\midrule
Few-shot, open-source & \\
\quad Mistral~\cite{mistral} & 78.3 \\
\quad Mistral + JSON & 83.0 \\
\quad Mistral + CoT & 81.5 \\
\quad Mistral + JSON+CoT & 82.4 \\
\midrule
Few-shot, proprietary & \\
\quad GPT-4~\cite{openai2023gpt4} & 87.6 \\
\quad GPT-4 + JSON & 86.4 \\
\quad GPT-4 + CoT & 87.6 \\
\quad GPT-4 + JSON+CoT & \bf 87.8 \\
\bottomrule
\end{tabular}
}
\caption{Accuracy on MedNLI under the 3-way classification setting. For 3-shot models, we provide one example from each class in the prompt. 
We organize results into 3 groups: fully supervised, 3-shot open-source, and 3-shot proprietary models.
}
\label{tab:NLI_3way}
\end{table}

\start{Experiments on claim recall}
We also compare different prompt styles for GPT-4 on the computation of claim recall.

We first randomly sample 20 (system output, reference claim) pairs from note generation where at least two evaluators make different entailment predictions. Then we ask 4 human annotators to annotate these examples.
We use the majority vote of the first three annotators’ answers as the ground truth and compute the accuracy of the fourth human annotator as well as the models. As shown in Table~\ref{tab:claim_acc20}, we can observe that the fourth annotator achieves 0.85 accuracy, indicating high agreement among annotators. Results also show that the agreement among humans is still stronger than the agreement between models and human annotators.

To obtain annotations for more examples, we additionally sample 100 (system output, reference claim) pairs where at least two evaluators disagree, and assign them to 5 human annotators. Namely, each human evaluator is assigned 20 examples. We consider human evaluations as the ground truth and compute the accuracy of each evaluator. The results of note generation are shown in Table~\ref{tab:claim_acc100}.
We can observe that both CoT and JSON improve the evaluation quality of claim recall (GPT-4).

\begin{table}[t]
\centering
\resizebox{0.85\linewidth}{!}{
\begin{tabular}{lc}
\toprule
\multicolumn{2}{l}{\bf Claim Recall Evaluation} \\
\multicolumn{2}{l}{(20 examples, three annotators per example)} \\
\bottomrule
\toprule
\bf Model
& \bf Accuracy \\
\midrule
TRUE & 60.0 \\
\midrule
GPT-4 (0-shot) & 40.0 \\
\quad + JSON & 50.0 \\
\quad + JSON + CoT & 80.0 \\
\midrule
GPT-4 (2-shot) & 50.0 \\
\quad + JSON & 55.0 \\
\quad + JSON + CoT & \bf 80.0 \\
\midrule
Human & \bf 85.0 \\
\bottomrule
\end{tabular}
}
\caption{Accuracy of each evaluator (including human) on claim recall evaluation on 20 sentences of ACI-BENCH. We take the majority vote of three annotators as the ground truth and compute the accuracy of another annotator (denoted as ``human'').}
\label{tab:claim_acc20}
\end{table}

\begin{table}[t]
\centering
\resizebox{0.8\linewidth}{!}{
\begin{tabular}{lc}
\toprule
\multicolumn{2}{l}{\bf Claim Recall Evaluation} \\
\multicolumn{2}{l}{(100 examples, one annotator per example)} \\
\bottomrule
\toprule
\bf Model
& \bf Accuracy \\
\midrule
TRUE & 58.0 \\
\midrule
GPT-4 (0-shot) & 52.0 \\
\quad + JSON & 64.0 \\
\quad + JSON + CoT & 65.0 \\
\midrule
GPT-4 (2-shot) & 48.0 \\
\quad + JSON & 62.0 \\
\quad + JSON + CoT & \bf 66.0 \\
\bottomrule
\end{tabular}
}
\caption{Accuracy of each evaluator on claim recall evaluation on ACI-BENCH. We select 100 sentences where at least two of these models have different predictions. We consider human annotations as the ground truth.}
\label{tab:claim_acc100}
\end{table}

\subsection{Examples of Different Prompt Styles}
\label{sec:evaluator_prompts}
We show the prompt styles ``CoT'' in \autoref{tab:prompt_aci_mistral} and ``JSON + CoT'' in \autoref{tab:prompt_aci_gpt4}. Based on the experiment results in \autoref{tab:NLI_new} and \S\ref{sec:prompt_style_experiments}, we use the prompt style ``CoT'' for the Mistral evaluator and use the prompt style ``JSON + CoT'' for the GPT-4 evaluator.

\subsection{Case Studies for Human Evaluation}
\autoref{tab:case_same} presents a case where the medical experts assign the same score to both outputs, but still prefer one over the other. The reason is that \textit{Output 2} contains the details about EKG, which supports ``the heart rate is normal'' with evidence. In comparison, \textit{Output 1} only mentions the heart rate is normal without revealing how the heart rate is examined.

\autoref{tab:case_rouge} presents an example where claim recall (GPT-4) disagrees with both ROUGE and MEDCON on preference over a pair of outputs. 

We explain the human judgment of each claim as follows:

\noindent \textit{- Output 1 supports Claim 1}: the output states ``patient is not in any distress''

\noindent \textit{- Output 1 supports Claim 2}: the output states ``Carotid: No appreciable carotid bruits''

\noindent \textit{- Output 1 supports Claim 3}: ``Lungs: Clear to auscultation bilaterally'' already means no wheezes, rales, or rhonchi.

\noindent \textit{- Output 1 supports Claim 4}: the output states ``Slight 2/6 systolic ejection murmur ''.

\noindent \textit{- Output 1 supports Claim 5}: the output states ``1+ edema in lower extremities''.

\noindent \textit{- Output 2 does not support Claim 1}: no information about distress.

\noindent \textit{- Output 2 does not support Claim 2}: no information about neck.

\noindent \textit{- Output 2 supports Claim 3}: ``clear lungs'' already means no wheezes, rales, or rhonchi. 

\noindent \textit{- Output 2 supports Claim 4}: the output states ``2/6 systolic ejection murmur''.

\noindent \textit{- Output 2 supports Claim 5}: the output states ``1+ pitting edema in bilateral lower extremities''.

\begin{table}[!t]
\centering

\resizebox{\columnwidth}{!}{
\begin{tabular}{p{10cm}}
\toprule
\colorbox{LightGrey}{\textit{\textbf{Reference subclaims}} \hskip17em} \\
1. The patient's carotid arteries do not have audible bruits. \\
2. The patient's lungs are clear to auscultation on both sides, with no wheezes, rales, or rhonchi. \\
3. The patient exhibits a slight 2/6 systolic ejection murmur in the cardiovascular exam, which is stable. \\
4. The patient has a normal heart rate. \\
5. The patient has trace lower extremity edema in both legs. \\

\midrule 

\colorbox{LightGrey}{\textit{\textbf{Output 1}} \hskip22em } \\
PHYSICAL EXAM: Heart rate is normal. On physical examination, there are no carotid bruits in the neck, but a slight 2/6 systolic ejection murmur is present on heart exam, which is stable. Lungs are clear, and there is trace lower extremity edema bilaterally. \\
\\
\textbf{Claim Recall (GPT-4)}: \textbf{100.00} \quad // Support all claims. \\
\textbf{Human Judgment}: \textbf{100.00} \quad // Support all claims. \\
\textbf{Preferred by Subjective Preference?}: \colorR{No.} \\

\midrule

\colorbox{LightGrey}{\textit{\textbf{Output 2} (Preferred by human)} \hskip13.2em} \\

OBJECTIVE EXAM: \\
- Carotid: No appreciable carotid bruits \\
- Heart: 2/6 systolic ejection murmur, stable from previous exams \\
- Lungs: Clear to auscultation bilaterally \\
- Extremities: Trace lower extremity edema bilaterally \\

EKG: Within normal limits \\
\\
\textbf{Claim Recall (GPT-4)}: \textbf{100.00} \quad // Support all claims. \\
\textbf{Human Judgment}: \textbf{100.00} \quad // Support all claims. \\
\textbf{Preferred by Subjective Preference?}: \colorG{Yes.} \quad // more details. \\

\bottomrule
\end{tabular}
}
\caption{
An example where human and claim recall (GPT-4) give the same entailment score for both outputs, but human subjectively prefers one of them. In this case, Output 2 also contains the EKG results, which is not presented in the reference but should be included.}
\label{tab:case_same}
\end{table}

\begin{table}[!t]
\centering

\resizebox{\columnwidth}{!}{
\begin{tabular}{p{10cm}}
\toprule

\colorbox{LightGrey}{\textit{\textbf{Reference subclaims}} \hskip17em} \\
1. The patient appears to be in no apparent distress. \\
2. No carotid bruits are present in the patient's neck. \\
3. The patient's lungs are clear upon auscultation, with no wheezes, rales, or rhonchi. \\
4. The patient has a grade 2/6 systolic ejection murmur. \\
5. There is 1+ edema in both lower extremities of the patient. \\

\midrule 

\colorbox{LightGrey}{\textit{\textbf{Output 1} (Preferred by GPT-4 and human)} \hskip8.6em } \\
OBJECTIVE EXAM: \\ 
Vital Signs: \\ 
- Blood pressure: within normal range \\ 
- Heart rate: within normal range \\ 
- Pulse ox: within normal range \\ 
Physical Exam: \\ 
- General: Patient is not in any distress \\ 
- Carotid: No appreciable carotid bruits \\ 
- Heart: Slight 2/6 systolic ejection murmur (previously noted) \\ 
- Lungs: Clear to auscultation bilaterally \\ 
- Extremities: 1+ edema in lower extremities \\ 
Labs: \\ 
- Blood glucose: 162 mg/dL (fasting) \\ 
- Hemoglobin A1c: 8\% \\
\\
\textbf{ROUGE-L Recall}: \colorR{27.66} \\
\textbf{MEDCON Recall}: \colorR{33.33} \\
\textbf{Claim Recall (GPT-4)}: \colorG{100.00} \quad // Support all claims. \\
\textbf{Human Judgment}: \colorG{100.00} \quad // Support all claims. \\

\midrule

\colorbox{LightGrey}{\textit{\textbf{Output 2} (Preferred by ROUGE-L and MEDCON)} \hskip5.5em} \\

PHYSICAL EXAMINATION \\ 
• Cardiovascular: 2/6 systolic ejection murmur, stable. \\ 
• Respiratory: Lungs clear to auscultation. \\ 
• Extremities: 1+ pitting edema in bilateral lower extremities. \\
\\
\textbf{ROUGE-L Recall}: \colorG{40.43} \\
\textbf{MEDCON Recall}: \colorG{75.00} \\
\textbf{Claim Recall (GPT-4)}: \colorR{60.00} \quad // Support Claim 3, 4, 5. \\
\textbf{Human Judgment}: \colorR{60.00} \quad // Support Claim 3, 4, 5. \\

\bottomrule
\end{tabular}
}
\caption{An example of disagreement between claim recall (GPT-4) and ROUGE/MEDCON on preferences over a pair of outputs. In this example, Output 1 has fewer medical terms overlapping with the reference, but covers more subclaims.}
\label{tab:case_rouge}
\end{table}

\begin{table*}[ht]
\centering
\resizebox{\textwidth}{!}{
\begin{tabular}{lcc}
\toprule
Task & Commonly-Used Metrics & Reference \\
\midrule
Clinical note generation & ROUGE, BERTScore, BLEURT, MEDCON & \citet{aci-bench,mts-dialog} \\
Radiology report summarization & ROUGE, BERTScore, BLEU, F1-RadGraph & \citet{radadapt,medpalmm} \\
Patient question summarization & ROUGE, BERTScore, BLEU, MEDCON & \citet{MeQSum,clinical-text-summ} \\
\bottomrule
\end{tabular}
}
\caption{Evaluated tasks and their commonly used metrics. We list prior works that apply the corresponding metrics under ``Reference''.}
\label{tab:metric_list}
\end{table*}

\begin{table*}[]
\centering
\resizebox{0.65\linewidth}{!}{
\begin{tabular}{lcc}
\toprule
\multirow{2}{*}{\bf Comparison} & \multicolumn{2}{c}{\bf \# Disagreements} \\
 & \bf O-Exam & \bf A \& P \\
\midrule
Claim Recall (GPT-4) \& ROUGE Recall           & 524 / 2460 & 635 / 2460 \\
Claim Recall (GPT-4) \& MEDCON Recall          & 251 / 2460 & 499 / 2460 \\
Claim Recall (GPT-4) \& Claim Recall (TRUE)    & 148 / 2460 & 268 / 2460 \\
Claim Recall (GPT-4) \& Claim Recall (Mistral) & 121 / 2460 & 309 / 2460 \\
\bottomrule
\end{tabular}
}
\caption{Number of disagreements between each two evaluation metrics among all 2460 pairs of generated outputs.}
\label{tab:disagreement_pairs}
\end{table*}

\begin{table*}[ht]
\centering
\resizebox{0.8\textwidth}{!}{
\begin{tabular}{lccc}
\toprule
\multicolumn{4}{l}{\textbf{Inter-annotator Agreement}} \\
\bottomrule
\toprule
\bf Comparison & \bf Section & \bf Spearman-$\rho$ & \bf Kendall-$\tau$ \\
\midrule
\multirow{2}{*}{Claim Recall (GPT-4) vs. Rouge Recall} & O-Exam & 0.744 & 0.641 \\
& A \& P & 0.881 & 0.767 \\
\midrule
\multirow{2}{*}{Claim Recall (GPT-4) vs. MEDCON Recall} & O-Exam & 0.688 & 0.594 \\
& A \& P & 0.785 & 0.667 \\
\midrule
\multirow{2}{*}{Claim Recall (GPT-4) vs. Claim Recall (TRUE)} & O-Exam & 0.765 & 0.644 \\
& A \& P & 0.686 & 0.554 \\
\midrule
\multirow{2}{*}{Claim Recall (GPT-4) vs. Claim Recall (Mistral)} & O-Exam & 0.690 & 0.561 \\
& A \& P & 0.861 & 0.709 \\
\bottomrule
\end{tabular}
}
\caption{Inter-annotator agreement in the human study (\S \ref{sec:human-preference}). We compare the correlation between one annotator and the average score given by the other two annotators.}
\label{tab:inter_annotator}
\end{table*}

\subsection{Experimental details}
\label{sec:exp_details}

\subsubsection{Datasets}
\label{sec:baseline}
\start{Clinical note generation experiments}
We experiment on the ACI-BENCH~\cite{aci-bench} dataset for note generation, which is a dataset of 207 pairs of dialogue and SOAP notes. We report results on the ``test1'' split, which contains 40 dialogue-note pairs.
The definition of ``SOAP note'' is a widely used method of documentation for healthcare providers \footnote{\href{https://www.ncbi.nlm.nih.gov/books/NBK482263/}{ncbi.nlm.nih.gov/books/NBK482263/}}.
The original task setup only evaluates the generated SOAP note against the reference. To assess attribution, our task setup additionally requires each sentence in the generated note to cite at least one conversational turn in the input dialogue that supports the sentence.

\start{Radiology report summarization experiments}
We conduct report summarization experiments on MIMIC-III~\cite{mimic}.
It contains 67K radiology reports spanning seven anatomies (head, abdomen, chest, spine, neck, sinus, and pelvis) and two modalities: magnetic resonance imaging (MRI) and computed tomography (CT).
In our experiments, we randomly sample 200 examples following the proportion of each Modality/Anatomy category (e.g., CT head, MR spine) for evaluation.
Similarly, we require each sentence to cite at least one sentence in the input.

\start{Patient question summarization experiments}
We conduct patient question summarization experiments on MeQSum~\cite{MeQSum}.
It contains (1) original patient health questions of varying verbosity and coherence selected from the U.S. National Library of Medicine, and (2) corresponding condensed questions created by three medical experts such
that the summary allows retrieving complete, correct answers to the original question without the potential for further condensation.
We experiment on the subset of the test set selected by the MEDIQA 2021 challenge~\cite{ben-abacha-etal-2021-overview}, which contains 100 examples and covers a wide range of topics and question types such as \emph{Treatment, Information, Side effects, Cause, Effect, Person-Organization, Diet-Lifestyle, Complications, Contraindications, Diagnosis, Usage, Interaction, Ingredients, Prognosis, Susceptibility, Transmission, and Toxicity}.

\begin{table*}[t]
\centering
\resizebox{\textwidth}{!}{
\begin{tabular}{p{20cm}}
\toprule
\colorbox{LightGrey}{\textit{\textbf{Prompt style}: JSON} \hskip44.2em } \\
\midrule
\underline{\textbf{Instruction:}} \\
Please act as an impartial judge and evaluate whether the clinical note provided by an AI assistant can fully entail each claim below. 
For each claim, please output '1' or '0' for each claim, where '1' means the claim can be fully entailed by the clinical note, and '0' means the claim contains information that cannot be entailed by the clinical note. \\
Generate the answer as a list of json dicts. Each dict should be in the format of \{'claim': the original claim, 'entailment prediction': 1 or 0, whether the claim can be entailed\}." \\
\\

\underline{\textbf{Example input:}} \\
\{ \\

\quad "clinical note": "PHYSICAL EXAM • Cardiovascular: 3/6 systolic ejection murmur, previously noted. • Extremities: 1+ pitting edema in lower extremities.", \\
\quad "claims": $[$ \\
\quad\quad "The patient's blood pressure is high.", \\
\quad\quad "The patient has a grade 3/6 systolic ejection murmur.", \\
\quad\quad "The patient exhibits 1+ pitting edema in both lower extremities." \\
\quad $]$ \\
\} \\
\\

\underline{\textbf{Example output:}} \\
$[$ \\
\quad\{ \\
\quad\quad "claim": "The patient's blood pressure is high.", \\
\quad\quad "entailment prediction": 0 \\
\quad\}, \\
\quad\{ \\
\quad\quad "claim": "The patient has a grade 3/6 systolic ejection murmur.", \\
\quad\quad "entailment prediction": 1 \\
\quad\}, \\
\quad\{ \\
\quad\quad  "claim": "The patient exhibits 1+ pitting edema in both lower extremities.", \\
\quad\quad "entailment prediction": 0 \\
\quad\} \\
$]$ \\
\bottomrule
\end{tabular}
}
\caption{Example of the prompt style ``JSON'' for claim recall and precision computation on ACI-BENCH. We format the input and output as a JSON dictionary.}
\label{tab:prompt_aci_json}
\end{table*}

\begin{table*}[t]
\centering
\resizebox{\textwidth}{!}{
\begin{tabular}{p{20cm}}
\toprule
\colorbox{LightGrey}{\textit{\textbf{Prompt style}: CoT} \hskip44.3em } \\
\midrule
\underline{\textbf{Instruction:}} \\
Please act as an impartial judge and evaluate whether the clinical note provided by an AI assistant can fully entail the claim below. Also generate an explanation for your answer. Please output '1' or '0' as your entailment prediction, where '1' means the claim can be fully entailed by the clinical note, and '0' means the claim contains information that cannot be entailed by the clinical note. Generate the answer in the following format: \\
explanation: the reason why the entailment prediction is made. \\ entailment prediction: 1 or 0, whether the claim can be entailed.
\\
\\
\underline{\textbf{Example input:}} \\
clinical note: PHYSICAL EXAM • Cardiovascular: 3/6 systolic ejection murmur, previously noted. • Extremities: 1+ pitting edema in lower extremities.\\

claim: The patient exhibits 1+ pitting edema in both lower extremities. \\
\\
\underline{\textbf{Example output:}} \\
explanation: the clinical note mentions 'Extremities: 1+ pitting edema', but does not specify whether it is in the upper or lower extremities. \\
entailment prediction: 0 \\
\bottomrule
\end{tabular}
}
\caption{Example of the prompt style ``CoT'' for claim recall and precision computation on ACI-BENCH. We prompt the model to generate an explanation for its prediction.}
\label{tab:prompt_aci_mistral}
\end{table*}

\begin{table*}[t]
\centering
\resizebox{\textwidth}{!}{
\begin{tabular}{p{20cm}}
\toprule
\colorbox{LightGrey}{\textit{\textbf{Prompt style}: JSON + CoT} \hskip41.5em } \\
\midrule
\underline{\textbf{Instruction:}} \\
Please act as an impartial judge and evaluate whether the clinical note provided by an AI assistant can fully entail each claim below. 
\textbf{Also generate an explanation for your answer.}
For each claim, please output '1' or '0' for each claim, where '1' means the claim can be fully entailed by the clinical note, and '0' means the claim contains information that cannot be entailed by the clinical note. \\
Generate the answer as a list of json dicts. Each dict should be in the format of \{'claim': the original claim, \textbf{'explanation': the reason why the entailment prediction is made,} 'entailment prediction': 1 or 0, whether the claim can be entailed\}." \\
\\

\underline{\textbf{Example input:}} \\
\{ \\

\quad "clinical note": "PHYSICAL EXAM • Cardiovascular: 3/6 systolic ejection murmur, previously noted. • Extremities: 1+ pitting edema in lower extremities.", \\
\quad "claims": $[$ \\
\quad\quad "The patient's blood pressure is high.", \\
\quad\quad "The patient has a grade 3/6 systolic ejection murmur.", \\
\quad\quad "The patient exhibits 1+ pitting edema in both lower extremities." \\
\quad $]$ \\
\} \\
\\

\underline{\textbf{Example output:}} \\
$[$ \\
\quad\{ \\
\quad\quad "claim": "The patient's blood pressure is high.", \\
\quad\quad \textbf{"explanation": "The clinical note does not mention anything about the blood pressure.",} \\
\quad\quad "entailment prediction": 0 \\
\quad\}, \\
\quad\{ \\
\quad\quad "claim": "The patient has a grade 3/6 systolic ejection murmur.", \\
\quad\quad \textbf{"explanation": "The PHYSICAL EXAM section mentions that 'Cardiovascular: 3/6 systolic ejection murmur', which supports the claim.",} \\
\quad\quad "entailment prediction": 1 \\
\quad\}, \\
\quad\{ \\
\quad\quad  "claim": "The patient exhibits 1+ pitting edema in both lower extremities.", \\
\quad\quad \textbf{"explanation": "The clinical note mentions 'Extremities: 1+ pitting edema', but does not specify whether it is in the upper or lower extremities.",} \\
\quad\quad "entailment prediction": 0 \\
\quad\} \\
$]$ \\
\bottomrule
\end{tabular}
}
\caption{Example of the prompt style ``JSON + CoT'' for claim recall and precision computation on ACI-BENCH. We format the input and output as a JSON dictionary and prompt the model to generate an explanation for its prediction. We highlight the difference between the ``JSON'' and ``JSON + CoT'' prompt styles in \textbf{bold}.}
\label{tab:prompt_aci_gpt4}
\end{table*}

\subsubsection{Examples for each Dataset}
\begin{table*}[!t]
\centering

\resizebox{0.95\textwidth}{!}{
\begin{tabular}{p{20cm}}
\toprule

\colorbox{LightGrey}{\textit{\textbf{Input}: Dialogue between the doctor and the patient} \hskip31em} \\

$[0]$ (doctor) hi, martha. how are you? \\ 
... \\
$[4]$ (doctor) so, martha, it's been a year since i've seen you. how are you doing? \\ 
$[5]$ (patient) i'm doing well. i've been traveling a lot recently since things have, have gotten a bit lighter. and i got my, my vaccine, so i feel safer about traveling. i've been doing a lot of hiking. uh, went to washington last weekend to hike in northern cascades, like around the mount baker area. \\ 
... \\ 
$[28]$ (doctor) so, i'm just gon na check out your heart and your lungs. and you know, let you know what i find, okay? \\  
$[29]$ (patient) okay. \\  
$[30]$ (doctor) okay. so, on your physical examination, you know, everything looks good. on your heart exam, i do appreciate a three out of six systolic ejection murmur, which i've heard in the past, okay? and on your lower extremities, i do appreciate one plus pitting edema, so you do have a little bit of fluid in your legs, okay? \\  
$[31]$ (patient) okay. \\ 
... \\
$[37]$ (doctor) i also wanna repeat another echocardiogram, okay? \\ 
$[38]$ (patient) okay. \\ 
... \\ 
$[46]$ (patient) can i take all my pills at the same time? \\ 
$[47]$ (doctor) yeah. \\ 
... \\

\midrule 

\colorbox{LightGrey}{\textit{\textbf{Reference}: Clinical note} \hskip41.8em } \\
\begin{mdframed}[backgroundcolor=yellow!20, linewidth=0pt]
CHIEF COMPLAINT \\ 
Annual exam $[4]$. \\ 
 \\ 
HISTORY OF PRESENT ILLNESS \\ 
Martha Collins is a 50-year-old female with a past medical history significant for congestive heart failure, depression, and hypertension who presents for her annual exam $[4]$. It has been a year since I last saw the patient $[4]$. \\ 
 \\ 
The patient has been traveling a lot recently since things have gotten a bit better $[5]$. She reports that she got her COVID-19 vaccine so she feels safer about traveling $[5]$. She has been doing a lot of hiking $[5]$. \\
... \\
\\
REVIEW OF SYSTEMS \\ 
• Ears, Nose, Mouth and Throat: Endorses nasal congestion from allergies $[22]$. \\ 
• Cardiovascular: Denies chest pain or dyspnea on exertion $[12]$$[13]$. \\ 
...
\end{mdframed}

\begin{mdframed}[backgroundcolor=green!20, linewidth=0pt]
PHYSICAL EXAMINATION \\ 
• Cardiovascular: Grade 3/6 systolic ejection murmur $[30]$. \\ 
1+ pitting edema of the bilateral lower extremities $[30]$. \\ 
 \\ 
VITALS REVIEWED \\ 
• Blood Pressure: Elevated $[26]$. \\ 
\end{mdframed}

\begin{mdframed}[backgroundcolor=cyan!20, linewidth=0pt]
RESULTS \\ 
Echocardiogram demonstrates decreased ejection fraction of 45\% $[32]$. Mitral regurgitation is present $[32]$. \\ 
Lipid panel: Elevated cholesterol $[33]$. \\ 
\end{mdframed}

\begin{mdframed}[backgroundcolor=blue!10, linewidth=0pt]
ASSESSMENT AND PLAN \\ 
Martha Collins is a 50-year-old female with a past medical history significant for congestive heart failure, depression, and hypertension who presents for her annual exam $[4]$. \\ 
 \\ 
Congestive heart failure. \\ 
• Medical Reasoning: She has been compliant with her medication and dietary modifications $[8]$$[9]$$[10]$$[11]$. Her previous year's echocardiogram demonstrated a reduced ejection fraction of 45\%, as well as some mitral regurgitation $[32]$. Her cholesterol levels were slightly elevated on her lipid panel from last year $[33]$. \\ 
• Additional Testing: We will order a repeat echocardiogram $[37]$$[38]$. We will also repeat a lipid panel this year $[33]$$[34]$. \\ 
...
\end{mdframed}
\\
\bottomrule
\end{tabular}
}
\caption{An example of note generation, where the input is the dialogue between the doctor and the patient, and the goal is to generate a clinical note based on the dialogue. We highlight the four sections in the note: \colorbox{LightYellow}{subjective}, \colorbox{LightGreen}{objective exam}, \colorbox{LightBlue}{objective results}, and \colorbox{LightPurple}{assessment and plan}.}
\label{tab:aci_example}
\end{table*}

\begin{table*}[t]
\centering

\resizebox{\textwidth}{!}{
\begin{tabular}{p{20cm}}
\toprule

\colorbox{LightGrey}{\textit{\textbf{Input}: The findings section of a radiology report} \hskip32.5em } \\
$[0]$ there are new areas of slow diffusion in left frontal and parietal lobes involving the precentral gyrus suggestive of acute infarcts. $[1]$ areas of slow diffusion are noted in left corona radiata, left thalamus, left parieto-occipital regions and splenium of corpus callosum which are unchanged since the prior study. $[2]$ areas of gliosis with flair and t2 hyperintensities are noted in bilateral occipital lobes which are sequelae of subacute/ chronic infarcts (left greater than right). $[3]$ there is no evidence of hemorrhagic transformation of infarcts. $[4]$ there is prominence of the cortical sulci, ventricular system and extra-axial csf spaces suggestive of generalized cerebral atrophy. $[5]$ mucosal thickening is noted in the sphenoid sinus and bilateral ethmoid air cells. $[6]$ orbits are unremarkable. $[7]$ there is partial opacification of bilateral mastoid air cells. \\
\\

\midrule 

\colorbox{LightGrey}{\textit{\textbf{Reference}: The impression section, which summarizes the findings section} \hskip22em } \\
1. new acute infarcts in left frontal and parietal lobes. $[0]$ \\
2. subacute infarcts in splenium of corpus callosum, left thalamus and left parietal and frontal white matter. $[0]$$[1]$ \\
3. sequelae of subacute/chronic infarcts in bilateral occipital lobes. $[2]$ \\
\\
\bottomrule
\end{tabular}
}
\caption{An example of report summarization, where the input is the findings section of a radiology report, and the goal is to generate an impression section that summarizes important results in the findings section.}
\label{tab:mimic_example}
\end{table*}

\begin{table*}[t]
\centering

\resizebox{\textwidth}{!}{
\begin{tabular}{p{20cm}}
\toprule

\colorbox{LightGrey}{\textit{\textbf{Input}: The original patient health questions} \hskip34em } \\
$[0]$ Hello, I have been dealing with trimethylaminuria since I was a child. $[1]$ I have done some of my own research and it looks like not much can be done for this condition. $[2]$ I do not have it all over my body. $[3]$ It’s only in my armpits. $[4]$ In the past I’ve gone to doctors and dermatologist they gave me no answers until I looked online today and finally found out what I have. $[5]$ I don’t know maybe I’m wrong. $[6]$ But this disease isn’t even consider common because no one has done anything about it. $[7]$ I’m sure they’re thousands of women with it... $[8]$ Can I be tested for it and help in some kind of way to finding a cure or something? $[9]$ What testing is done for this? $[10]$ And where? $[11]$ Thank you \\
\\

\midrule 

\colorbox{LightGrey}{\textit{\textbf{Output}: The summarized question} \hskip38em } \\
How can I get tested and treated for trimethylaminuria? $[0][8][9][10]$ \\
\\
\bottomrule
\end{tabular}
}
\caption{An example of question summarization, where the input is the patient question of varying verbosity and coherence, and the goal is to summarize the input into a short question that allows retrieving complete, correct answers to the original question.}
\label{tab:meqsum_example}
\end{table*}

\autoref{tab:aci_example} is an example of clinical note generation under the full-note generation setting. The input is the dialogue between the doctor and the patient and the target output is the full clinical note based on the dialogue. We highlight the four sections in the reference clinical note, which is divided automatically by the name of each paragraph (e.g., ``REVIEW OF SYSTEMS'' belongs to the subjective section).
As for per-section generation, the input is the same and the target output contains only one of the four sections.

\autoref{tab:mimic_example} is an example of radiology report summarization. The input is the findings section of the radiology report, containing experiment results and findings. The target output is the impression section of the report, which should summarize the important information in the findings section.

\autoref{tab:meqsum_example} shows an example of patient question summarization. The input is the patient question of varying verbosity and coherence, and the goal is to summarize the input into a short question that allows retrieving complete, correct answers to the original question.

\subsubsection{Evaluation Metrics}
To evaluate note generation, we compare \modelname with the following metrics that are commonly used in the existing research~\cite{aci-bench}:
\textbf{ROUGE}~\cite{rouge} computes the overlap of n-grams.
\textbf{BERTScore}~\cite{bertscore} compares the embeddings of matched tokens in the output and reference.
\textbf{BLEURT}~\cite{bleurt} trains a model to compute output-reference similarity.
\textbf{MEDCON}~\cite{aci-bench} computes the F1-score of the UMLS concepts in the output and the reference.

To evaluate report summarization, we follow previous work \cite{radadapt} and evaluate ROUGE, BERTScore, BLEU, and F1-RadGraph \cite{radgraph}, where BLEU evaluates the overlap of 1- to 4-grams, and F1-RadGraph computes the F1 score of a predefined set of entities and relations present in radiology reports.

As for question summarization, we evaluate ROUGE, BERTScore, BLEU, and MEDCON following existing words~\cite{clinical-text-summ,MeQSum}.

\subsubsection{Evaluated Methods}
We evaluate both open-source models and proprietary models on note generation.
For open-source models, we choose the best models reported by \newcite{aci-bench}: BART fine-tuned on SAMSum \cite{samsum} (denoted as BART + SAMSum FT) and BioBART \cite{yuan-etal-2022-biobart}.
Since these models are not capable of generating citations, we do not report their citation metrics.
For proprietary models, we experiment on GPT-3.5-turbo and GPT-4 with zero-shot and few-shot prompting. 

As for report summarization, we compare GPT-3.5-turbo and GPT-4 with zero-shot and few-shot prompting. For few-shot prompts, we sample the same number of examples (1 or 2) from each of the six Modality/Anatomy categories in MIMIC that have a train set. Namely, we experiment on 0-shot, 6-shot, and 12-shot prompting.

Similarly, to evaluate question summarization, we compare 3.5-turbo and GPT-4 with 0-shot, 1-shot, and 2-shot prompting. The few-shot examples are selected from the validation set of MeQSum.

\subsubsection{Human Evaluation Details}
\label{sec:annotation_details}
We only annotate the disagreement between two metrics to manage the required amount of human annotation. As shown in \autoref{tab:disagreement_pairs}, the metrics have agreed preferences in most of the cases. For instance, if we randomly sample a set of output pairs for the objective exam section, GPT-4 and TRUE disagree in only 148 / 2640 = 5.6\% of the output pairs. GPT-4 and Mistral disagree in only 121 / 2640 = 4.6\% of the output pairs. As a result, randomly sampling from all output pairs would be inefficient, and we only focus on the disagreement.

We focus on completeness evaluation because in medical scenarios, omission errors (i.e., important information is missed or excluded) are more critical than commission errors~\cite{sin-omission} or hallucinations~\cite{med-omit}, in which information is fabricated and erroneously included. In the most common setting where we have human experts in the loop, detection of hallucinations is a much easier task as it can rely on comparisons against cited input or external sources. In contrast, detecting erroneous omissions is especially challenging as they are by definition not present in a system output, yet can mislead a reader by incorrectly portraying the source document.

We have five medical experts from two countries participating in our human evaluation, as introduced in \S \ref{sec:human-preference}.
All of them are researchers in biomedical machine learning. We held a 1-hour meeting to briefly introduce our work and explain the purpose and setting of the human study.

\begin{figure*}[!ht]
    \centering
    \includegraphics[width=\linewidth]{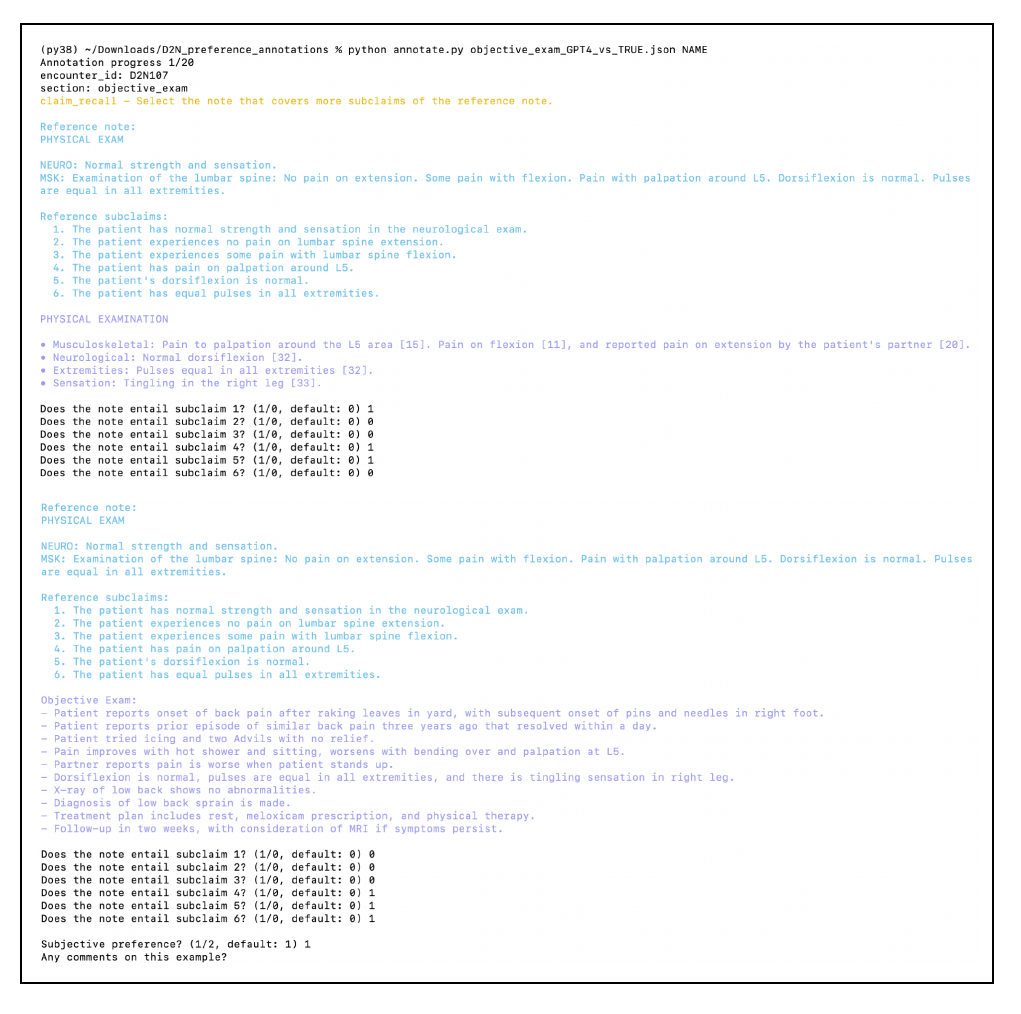}
\upv
\caption{The screenshot of the interface of our human study. Each annotator is asked to decide (1) whether each claim is fully supported by the two notes, and (2) which note they subjectively think is more complete (i.e., covers more information in the reference note). They are optionally asked to provide their comments for each example.}
\downv
\label{fig:annotation_screenshot}
\end{figure*}

We provide the instructions for the annotators in \autoref{tab:instruction} and provide a screenshot of the interface in \autoref{fig:annotation_screenshot}. We provide the annotators with the reference note, the extracted subclaims of the reference note, and two notes generated by two models. We do not tell the annotators which model generates which note. The annotators are asked to judge (1) whether each claim is fully supported by the two notes, and (2) which note they subjectively think is more complete (i.e., covers more information in the reference note).
To better understand the judgments of the annotators, they are free to leave any comments or thoughts when annotating the example, including but not limited to the explanations of their subjective preferences, and whether the claims extracted from the reference note are accurate and complete.

\begin{table*}[]
\centering
\resizebox{0.9\textwidth}{!}{
\begin{tabular}{p{\linewidth}}
\toprule
\bf Instructions for the annotators \\
\midrule
We are working on a project on medical text evaluation, and we need your help to evaluate the quality of clinical notes generated by different models. In this study, you will be presented with 20 (or 10) examples. Each example contains 2 clinical notes generated by different models, the reference clinical note written by human, and a list of subclaims we extracted from the reference clinical note. \\

\\

Your tasks are as follows: \\
(1) Judge whether the information in each subclaim is fully covered by each of the generated clinical notes. \\
(2) Judge which clinical note is more complete. Namely, which clinical note better captures the important information in the reference clinical note. \\
\bottomrule
\end{tabular}
}
\caption{Instructions for the annotators in our human study. We present the instructions in a meeting in which we briefly introduced them to the task of note generation, our project, and the purpose of the human study. }
\label{tab:instruction}
\end{table*}

\subsection{Detailed Evaluation Results}
For \modelname computed with GPT-4, we provide the detailed evaluation results of note generation in \autoref{tab:ACI_section}, which corresponds to \autoref{fig:aci_performance}.
Evaluation results on report summarization are shown in \autoref{tab:MIMIC_orig}, which corresponds to \autoref{fig:mimic_performance}.
Evaluation results on question summarization are in \autoref{tab:meqsum}, which corresponds to \autoref{fig:meqsum_performance}.

\Cref{tab:ACI_TRUE,tab:MIMIC_TRUE,tab:meqsum_TRUE} show the results evaluated by \modelname computed with TRUE.

\Cref{tab:ACI_Mistral_combine,tab:ACI_Mistral,tab:MIMIC_Mistral,tab:meqsum_Mistral} show the results evaluated by \modelname computed with Mistral.

\subsubsection{Generation With or Without Citations}
We compare the performance with or without asking the model to generate citations. As shown in \autoref{tab:MIMIC_citation_or_not}, there are no significant differences in the performances of generating with or without citations.

\begin{table*}[t]
\centering
\resizebox{0.6\linewidth}{!}{
\begin{tabular}{lcccc}
\toprule
\multicolumn{5}{l}{\textbf{Evaluation Methods:} Rouge, MEDCON} \\
\bottomrule
\toprule
\bf Model 
& \bf Rouge-1 & \bf Rouge-2 & \bf Rouge-L & \bf MEDCON \\
\midrule
BART + SAMSum FT (full) & 40.87 & 18.96 & 34.60 & 41.69 \\
BART + SAMSum FT (section) & 53.46 & 25.08 & 48.62 & 48.37 \\
BioBART (full) & 39.09 & 17.24 & 33.19 & 43.05 \\
BioBART (section) & 49.53 & 22.47 & 44.92 & 43.21 \\
\midrule
GPT-3.5-turbo (full, 0-shot) & 48.25 & 20.54 & 43.78 & 56.82 \\
GPT-3.5-turbo (section, 0-shot) & 47.61 & 20.80 & 43.65 & 57.48 \\
GPT-4 (full, 0-shot) & 48.74 & 22.93 & 45.26 & 59.49 \\
GPT-4 (section, 0-shot) & 51.16 & 23.33 & 46.98 & 60.01 \\
\midrule
GPT-3.5-turbo-16k (full, 1-shot) & 53.08 & 24.16 & 48.71 & 57.96 \\
GPT-3.5-turbo-16k (section, 1-shot) & 50.56 & 23.85 & 47.21 & 57.52 \\
GPT-4 (full, 1-shot) & 56.34 & 27.04 & 51.62 & 62.84 \\
GPT-4 (section, 1-shot) & 56.99 & 28.10 & 52.74 & 62.07 \\
\midrule
GPT-3.5-turbo-16k (full, 2-shot) & 56.50 & 27.38 & 52.27 & 60.13 \\
GPT-3.5-turbo-16k (section, 2-shot) & 51.99 & 25.52 & 48.22 & 58.80 \\
GPT-4-32k (full, 2-shot) & \bf 58.50 & 29.44 & \bf 54.34 & \bf 63.00 \\
GPT-4-32k (section, 2-shot) & 58.25 & \bf 29.73 & 53.96 & 61.95 \\
\bottomrule
\end{tabular}
}
\caption{Note generation results on ACI-BENCH-test1 evaluated with existing metrics. We compute each metric over the full note.}
\label{tab:ACI_existing}
\end{table*}

\begin{table*}[t]
\centering
\resizebox{0.7\linewidth}{!}{
\begin{tabular}{lcccc}
\toprule
\multicolumn{5}{l}{\textbf{Evaluation Method:} \modelname computed with GPT-4} \\
\bottomrule
\toprule
\bf Model & \bf Claim Recall & \bf Claim Prec & \bf Citation Recall & \bf Citation Prec \\
\midrule
BART + SAMSum FT (full) & 20.30 & 48.11 & -- & -- \\
BART + SAMSum FT (section) & 33.59 & 30.38 & -- & -- \\
BioBART (full) & 16.53 & 42.43 & -- & -- \\
BioBART (section) & 29.37 & 26.10 & -- & -- \\
\midrule
GPT-3.5-turbo (full, 0-shot) & 46.80 & 69.74 & 53.99 & 42.16 \\
GPT-3.5-turbo (section, 0-shot) & 56.28 & 29.18 & 58.77 & 49.31 \\
GPT-4 (full, 0-shot) & 48.31 & \bf 75.67 & 68.74 & \bf 66.07 \\
GPT-4 (section, 0-shot) & 60.24 & 36.73 & 61.13 & 58.47 \\
\midrule
GPT-3.5-turbo-16k (full, 1-shot) & 49.96 & 71.94 & 65.32 & 63.12 \\
GPT-3.5-turbo-16k (section, 1-shot) & 59.13 & 39.29 & 62.63 & 60.94 \\
GPT-4 (full, 1-shot) & 58.26 & 73.90 & 69.94 & 65.72 \\
GPT-4 (section, 1-shot) & 67.69 & 52.47 & \bf 70.81 & 65.75 \\
\midrule
GPT-3.5-turbo-16k (full, 2-shot) & 54.92 & 68.89 & 63.54 & 61.83 \\
GPT-3.5-turbo-16k (section, 2-shot) & 62.44 & 42.91 & 65.79 & 64.09 \\
GPT-4-32k (full, 2-shot) & 59.07 & 69.20 & 68.74 & 64.17 \\
GPT-4-32k (section, 2-shot) & \bf 69.88 & 54.92 & 67.90 & 63.01 \\
\bottomrule
\end{tabular}
}
\caption{Note generation results on ACI-BENCH-test1 evaluated with \modelname computed with GPT-4. We compute each metric over the full note.}
\label{tab:ACI_combine}
\end{table*}

\begin{table*}[t]
\centering
\resizebox{0.7\linewidth}{!}{
\begin{tabular}{lcccc}
\toprule
\multicolumn{5}{l}{\textbf{Evaluation Method:} \modelname computed with Mistral} \\
\bottomrule
\toprule
\bf Model & \bf Claim Recall & \bf Claim Prec & \bf Citation Recall & \bf Citation Prec \\
\midrule
BART + SAMSum FT (full) & 24.76 & 52.42 & -- & -- \\
BART + SAMSum FT (section) & 43.03 & 36.08 & -- & -- \\
BioBART (full) & 18.77 & 47.63 & -- & -- \\
BioBART (section) & 36.84 & 32.95 & -- & -- \\
\midrule
GPT-3.5-turbo (full, 0-shot) & 60.97 & 75.79 & 91.31 & 80.03 \\
GPT-3.5-turbo (section, 0-shot) & 67.30 & 36.44 & 83.34 & 64.20 \\
GPT-4 (full, 0-shot) & 62.13 & \bf 79.23 & \bf 94.00 & \bf 89.65 \\
GPT-4 (section, 0-shot) & 72.31 & 42.33 & 89.47 & 80.58 \\
\midrule
GPT-3.5-turbo-16k (full, 1-shot) & 61.50 & 75.58 & 85.68 & 82.79 \\
GPT-3.5-turbo-16k (section, 1-shot) & 65.20 & 46.91 & 88.64 & 80.19 \\
GPT-4 (full, 1-shot) & 69.53 & 78.56 & 89.01 & 82.90 \\
GPT-4 (section, 1-shot) & 72.59 & 57.32 & 91.34 & 82.18 \\
\midrule
GPT-3.5-turbo-16k (full, 2-shot) & 66.05 & 74.04 & 84.60 & 79.10 \\
GPT-3.5-turbo-16k (section, 2-shot) & 66.30 & 50.82 & 88.32 & 81.90 \\
GPT-4-32k (full, 2-shot) & 67.89 & 73.17 & 87.86 & 80.06 \\
GPT-4-32k (section, 2-shot) & \bf 73.96 & 59.24 & 86.15 & 77.57 \\
\bottomrule
\end{tabular}
}
\caption{Note generation results on ACI-BENCH-test1 evaluated with \modelname computed with Mistral. We compute each metric over the full note.}
\label{tab:ACI_Mistral_combine}
\end{table*}

\begin{table*}[t]
\centering
\resizebox{0.7\linewidth}{!}{
\begin{tabular}{lcccc}
\toprule
\multicolumn{5}{l}{\textbf{Evaluation Method:} \modelname computed with TRUE} \\
\bottomrule
\toprule
\bf Model & \bf Claim Recall & \bf Claim Prec & \bf Citation Recall & \bf Citation Prec \\
\midrule
BART + SAMSum FT (full) & 19.95 & 47.62 & -- & -- \\
BART + SAMSum FT (section) & 32.23 & 29.42 & -- & -- \\
BioBART (full) & 15.74 & 40.45 & -- & -- \\
BioBART (section) & 27.84 & 24.05 & -- & -- \\
\midrule
GPT-3.5-turbo (full, 0-shot) & 41.76 & 63.25 & 68.27 & 55.52 \\
GPT-3.5-turbo (section, 0-shot) & 53.54 & 27.14 & 56.20 & 34.52 \\
GPT-4 (full, 0-shot) & 45.21 & \bf 72.61 & \bf 72.41 & \bf 64.53 \\
GPT-4 (section, 0-shot) & 57.79 & 34.53 & 49.38 & 38.59 \\
\midrule
GPT-3.5-turbo-16k (full, 1-shot) & 47.40 & 66.28 & 56.51 & 52.23 \\
GPT-3.5-turbo-16k (section, 1-shot) & 56.39 & 37.57 & 58.45 & 48.03 \\
GPT-4 (full, 1-shot) & 56.61 & 69.49 & 60.95 & 51.92 \\
GPT-4 (section, 1-shot) & 67.13 & 49.81 & 60.99 & 49.13 \\
\midrule
GPT-3.5-turbo-16k (full, 2-shot) & 53.04 & 65.19 & 54.99 & 49.50 \\
GPT-3.5-turbo-16k (section, 2-shot) & 60.71 & 41.00 & 59.59 & 52.18 \\
GPT-4-32k (full, 2-shot) & 57.90 & 63.83 & 54.41 & 46.42 \\
GPT-4-32k (section, 2-shot) & \bf 67.63 & 50.78 & 55.33 & 45.99 \\
\bottomrule
\end{tabular}
}
\caption{Note generation results on ACI-BENCH-test1 evaluated with \modelname computed with TRUE. We compute each metric over the full note.}
\label{tab:ACI_TRUE_combine}
\end{table*}

\begin{table*}[!ht]
\centering
\resizebox{0.6\linewidth}{!}{
\begin{tabular}{lcccccc}
\toprule
\multicolumn{5}{l}{\textbf{Evaluation Methods:} Rouge, BERTScore, BLEURT, MEDCON} \\
\bottomrule
\toprule
\multicolumn{7}{c}{\bf Subjective} \\
\midrule
\bf Model 
& \bf Rouge-1 & \bf Rouge-2 & \bf Rouge-L & \bf BERTScore-F1 & \bf BLEURT & \bf MEDCON \\
\midrule
Reported GPT-3.5-turbo (full, 0-shot) & 32.70 & 14.05 & 22.69 & 65.14 & 39.48 & 38.21 \\
Reported GPT-4 (full, 0-shot) & 41.20 & 19.02 & 26.56 & 63.34 & 43.18 & 44.25 \\
\midrule
BART + SAMSum FT (full) & 46.33 & 25.52 & 29.88 & 68.68 & 45.00 & 43.02 \\ 
BART + SAMSum FT (section) & 52.44 & \bf 30.44 & 35.83 & 72.41 & 44.51 & 47.68 \\ 
BioBART (full) & 45.79 & 23.65 & 28.96 & 68.49 & 41.09 & 41.10 \\ 
BioBART (section) & 46.29 & 25.99 & 32.43 & 70.30 & 42.98 & 41.21 \\
\midrule
GPT-3.5-turbo (full, 0-shot) & 33.56 & 15.15 & 23.54 & 63.73 & 42.48 & 40.81 \\
GPT-3.5-turbo (section, 0-shot) & 36.30 & 13.27 & 20.30 & 59.69 & 43.55 & 35.95 \\
GPT-4 (full, 0-shot) & 35.09 & 16.49 & 25.56 & 65.12 & 42.11 & 41.15 \\
GPT-4 (section, 0-shot) & 43.32 & 19.03 & 27.04 & 63.70 & 46.30 & 48.54 \\
\midrule
GPT-3.5-turbo-16k (full, 1-shot) & 43.02 & 21.97 & 31.15 & 70.37 & 45.62 & 48.50 \\
GPT-3.5-turbo-16k (section, 1-shot) & 40.95 & 18.15 & 29.21 & 62.37 & 43.57 & 38.02 \\
GPT-4 (full, 1-shot) & 47.63 & 24.71 & 33.07 & 71.51 & 46.58 & \bf 50.96 \\
GPT-4 (section, 1-shot) & 48.16 & 23.23 & 34.08 & 65.04 & 47.86 & 44.97 \\
\midrule
GPT-3.5-turbo-16k (full, 2-shot) & 50.44 & 27.18 & 35.55 & 73.68 & \bf 48.62 & 49.18 \\
GPT-3.5-turbo-16k (section, 2-shot) & 40.95 & 18.76 & 29.80 & 63.27 & 44.52 & 35.92 \\
GPT-4-32k (full, 2-shot) & \bf 53.45 & 30.12 & \bf 38.68 & \bf 75.44 & 48.03 & 49.65 \\
GPT-4-32k (section, 2-shot) & 49.58 & 24.71 & 34.86 & 65.81 & 47.55 & 45.95 \\
\bottomrule
\toprule
\multicolumn{7}{c}{\bf Objective-Exam} \\
\midrule
\bf Model 
& \bf Rouge-1 & \bf Rouge-2 & \bf Rouge-L & \bf BERTScore-F1 & \bf BLEURT & \bf MEDCON \\
\midrule
Reported GPT-3.5-turbo (full, 0-shot) & 49.44 & 27.29 & 38.60 & 71.39 & 49.39 & 48.95 \\
Reported GPT-4 (full, 0-shot) & 50.11 & 28.20 & 40.43 & 71.79 & 51.11 & 42.59 \\
\midrule
BART + SAMSum FT (full) & 6.22 & 3.74 & 5.21 & 44.33 & 14.83 & 4.14 \\ 
BART + SAMSum FT (section) & 47.73 & 29.51 & 36.98 & 73.41 & 42.86 & 35.91 \\ 
BioBART (full) & 2.57 & 1.04 & 1.68 & 42.10 & 12.40 & 1.22 \\ 
BioBART (section) & 42.51 & 26.15 & 32.19 & 71.57 & 42.18 & 29.55 \\
\midrule
GPT-3.5-turbo (full, 0-shot) & 51.19 & 31.04 & 42.10 & 71.23 & 52.15 & 49.02 \\
GPT-3.5-turbo (section, 0-shot) & 32.21 & 17.87 & 26.44 & 60.11 & 45.57 & 31.60 \\
GPT-4 (full, 0-shot) & 62.41 & 40.44 & 52.82 & 77.04 & 55.40 & 55.69 \\
GPT-4 (section, 0-shot) & 39.58 & 22.91 & 35.11 & 62.70 & 45.55 & 38.33 \\
\midrule
GPT-3.5-turbo-16k (full, 1-shot) & 55.67 & 32.25 & 45.85 & 76.74 & 51.77 & 44.25 \\
GPT-3.5-turbo-16k (section, 1-shot) & 39.07 & 22.02 & 34.52 & 60.45 & 37.69 & 36.29 \\
GPT-4 (full, 1-shot) & 63.18 & 40.99 & 53.14 & 80.52 & \bf 55.74 & \bf 55.04 \\
GPT-4 (section, 1-shot) & 52.45 & 30.10 & 46.81 & 66.45 & 44.42 & 49.54 \\
\midrule
GPT-3.5-turbo-16k (full, 2-shot) & 61.13 & 37.93 & 51.84 & 79.40 & 52.18 & 48.14 \\
GPT-3.5-turbo-16k (section, 2-shot) & 46.46 & 27.29 & 42.44 & 64.02 & 41.24 & 44.97 \\
GPT-4-32k (full, 2-shot) & \bf 65.98 & \bf 42.61 & \bf 53.65 & \bf 81.43 & 51.50 & 53.80 \\
GPT-4-32k (section, 2-shot) & 55.58 & 32.45 & 49.69 & 68.54 & 47.29 & 50.50 \\
\bottomrule
\toprule
\multicolumn{7}{c}{\bf Objective-Results} \\
\midrule
\bf Model 
& \bf Rouge-1 & \bf Rouge-2 & \bf Rouge-L & \bf BERTScore-F1 & \bf BLEURT & \bf MEDCON \\
\midrule
Reported GPT-3.5-turbo (full, 0-shot) & 34.50 & 17.75 & 30.84 & 66.68 & 48.51 & 22.28 \\
Reported GPT-4 (full, 0-shot) & 37.65 & 19.94 & 35.73 & 68.33 & 48.50 & 26.73 \\
\midrule
BART + SAMSum FT (full) & 20.79 & 0.46 & 20.67 & 54.54 & 28.35 & 0.77 \\ 
BART + SAMSum FT (section) & 29.45 & 18.01 & 26.63 & 66.43 & 40.76 & 20.17 \\ 
BioBART (full) & 17.50 & 0.00 & 17.50 & 52.44 & 25.35 & 0.00 \\ 
BioBART (section) & 35.38 & 14.33 & 32.79 & 68.40 & 47.64 & 15.69 \\
\midrule
GPT-3.5-turbo (full, 0-shot) & 36.01 & 20.57 & 33.45 & 66.06 & 51.00 & 22.27 \\
GPT-3.5-turbo (section, 0-shot) & 32.88 & 5.68 & 31.81 & 63.49 & 54.42 & 6.94 \\
GPT-4 (full, 0-shot) & 45.81 & 28.78 & 43.76 & 70.62 & 54.60 & 36.23 \\
GPT-4 (section, 0-shot) & 34.11 & 6.46 & 32.62 & 62.94 & 52.69 & 7.05 \\
\midrule
GPT-3.5-turbo-16k (full, 1-shot) & 47.94 & 30.74 & 46.36 & 77.69 & 58.28 & \bf 36.64 \\
GPT-3.5-turbo-16k (section, 1-shot) & 40.26 & 9.07 & 39.03 & 65.50 & 56.38 & 11.35 \\
GPT-4 (full, 1-shot) & \bf 64.59 & \bf 33.67 & \bf 63.26 & \bf 86.09 & \bf 68.34 & 37.27 \\
GPT-4 (section, 1-shot) & 46.13 & 12.34 & 43.86 & 67.85 & 58.69 & 15.71 \\
\midrule
GPT-3.5-turbo-16k (full, 2-shot) & 47.87 & 33.08 & 46.82 & 77.34 & 56.74 & 34.62 \\
GPT-3.5-turbo-16k (section, 2-shot) & 44.94 & 12.90 & 43.45 & 67.92 & 61.29 & 13.97 \\
GPT-4-32k (full, 2-shot) & 61.87 & 32.35 & 60.66 & 84.54 & 65.40 & 32.95 \\
GPT-4-32k (section, 2-shot) & 45.63 & 12.55 & 44.10 & 68.05 & 58.58 & 15.02 \\
\bottomrule
\toprule
\multicolumn{7}{c}{\bf Assessment and Plan} \\
\midrule
\bf Model 
& \bf Rouge-1 & \bf Rouge-2 & \bf Rouge-L & \bf BERTScore-F1 & \bf BLEURT & \bf MEDCON \\
\midrule
Reported GPT-3.5-turbo (full, 0-shot) & 36.43 & 12.50 & 23.32 & 63.56 & 48.21 & 43.71 \\
Reported GPT-4 (full, 0-shot) & 38.16 & 14.12 & 24.90 & 64.26 & 49.41 & 42.36 \\
\midrule
BART + SAMSum FT (full) & 1.52 & 0.49 & 0.87 & 35.38 & 19.80 & 1.00 \\ 
BART + SAMSum FT (section) & 43.89 & 21.37 & 27.56 & 68.09 & 41.95 & 31.65 \\ 
BioBART (full) & 0.00 & 0.00 & 0.00 & 0.00 & 29.05 & 0.00 \\ 
BioBART (section) & 42.44 & 19.44 & 26.42 & 67.57 & 43.88 & 31.07 \\
\midrule
GPT-3.5-turbo (full, 0-shot) & 37.42 & 14.20 & 25.14 & 64.38 & 49.79 & 47.63 \\
GPT-3.5-turbo (section, 0-shot) & 38.17 & 14.10 & 26.80 & 62.41 & 51.16 & 43.75 \\
GPT-4 (full, 0-shot) & 39.25 & 16.12 & 27.71 & 68.13 & 50.91 & 47.03 \\
GPT-4 (section, 0-shot) & 41.68 & 16.52 & 30.14 & 63.47 & \bf 52.44 & 43.20 \\
\midrule
GPT-3.5-turbo-16k (full, 1-shot) & 48.20 & 23.16 & 33.58 & 71.79 & 51.31 & 50.00 \\
GPT-3.5-turbo-16k (section, 1-shot) & 43.93 & 17.98 & 31.81 & 63.98 & 48.44 & 43.61 \\
GPT-4 (full, 1-shot) & 49.32 & 23.50 & 33.74 & 72.14 & 50.26 & 50.07 \\
GPT-4 (section, 1-shot) & 46.48 & 20.38 & 34.18 & 65.57 & 47.39 & 44.80 \\
\midrule
GPT-3.5-turbo-16k (full, 2-shot) & 52.18 & 26.99 & 35.83 & 73.25 & 48.47 & 47.82 \\
GPT-3.5-turbo-16k (section, 2-shot) & 44.68 & 19.38 & 32.34 & 64.68 & 47.19 & 43.99 \\
GPT-4-32k (full, 2-shot) & \bf 52.87 & \bf 28.10 & \bf 36.28 & \bf 73.78 & 49.39 & \bf 51.49 \\
GPT-4-32k (section, 2-shot) & 46.64 & 22.11 & 34.36 & 65.72 & 47.54 & 46.13 \\
\bottomrule
\end{tabular}
}
\caption{Note generation results of each section on ACI-BENCH-test1, evaluated by existing metrics.}
\label{tab:ACI_existing_section}
\end{table*}

\begin{table*}[!ht]
\centering
\resizebox{0.55\linewidth}{!}{
\begin{tabular}{lcccc}
\toprule
\multicolumn{5}{l}{\textbf{Evaluation Method:} \modelname computed with GPT-4} \\
\bottomrule
\toprule
\multicolumn{5}{c}{\bf Subjective} \\
\midrule
\bf Model 
& \bf Claim Recall & \bf Claim Prec & \bf Citation Recall & \bf Citation Prec \\
\midrule
BART + SAMSum FT (full) & 46.98 & 49.50 & -- & -- \\ 
BART + SAMSum FT (section) & 42.94 & 55.90 & -- & -- \\ 
BioBART (full) & 39.92 & 44.00 & -- & -- \\ 
BioBART (section) & 33.69 & 50.68 & -- & -- \\
\midrule
GPT-3.5-turbo (full, 0-shot) & 32.30 & 72.84 & 57.42 & 40.21 \\
GPT-3.5-turbo (section, 0-shot) & 48.82 & 37.90 & 56.29 & 38.71 \\
GPT-4 (full, 0-shot)& 32.22 & 75.85 & 76.21 & 68.90 \\
GPT-4 (section, 0-shot) & 51.02 & 66.36 & 71.96 & 65.77 \\
\midrule
GPT-3.5-turbo-16k (full, 1-shot) & 41.06 & 71.63 & 69.65 & 64.04 \\
GPT-3.5-turbo-16k (section, 1-shot) & 58.55 & 44.76 & 68.72 & 54.94 \\
GPT-4 (full, 1-shot) & 47.39 & \bf 75.53 & \bf 79.57 & \bf 69.91 \\
GPT-4 (section, 1-shot) & 63.76 & 64.50 & 74.05 & 65.96 \\
\midrule
GPT-3.5-turbo-16k (full, 2-shot) & 50.65 & 67.65 & 71.16 & 63.85 \\
GPT-3.5-turbo-16k (section, 2-shot) & 62.47 & 41.71 & 67.19 & 63.74 \\
GPT-4-32k (full, 2-shot) & 52.42 & 72.29 & 75.78 & 68.13 \\
GPT-4-32k (section, 2-shot) & \bf 63.91 & 67.57 & 73.77 & 66.06 \\
\bottomrule
\toprule
\multicolumn{5}{c}{\bf Objective-Exam} \\
\midrule
\bf Model 
& \bf Claim Recall & \bf Claim Prec & \bf Citation Recall & \bf Citation Prec \\
\midrule
BART + SAMSum FT (full) & 1.71 & 21.94 & -- & -- \\ 
BART + SAMSum FT (section) & 32.48 & 20.65 & -- & -- \\ 
BioBART (full) & 1.04 & 15.83 & -- & -- \\ 
BioBART (section) & 34.08 & 16.41 & -- & -- \\
\midrule
GPT-3.5-turbo (full, 0-shot) & 61.30 & 68.88 & 57.25 & 47.07 \\
GPT-3.5-turbo (section, 0-shot)& 62.27 & 22.33 & 63.90 & 52.64 \\
GPT-4 (full, 0-shot) & 64.28 & 69.22 & 62.00 & 59.38 \\
GPT-4 (section, 0-shot) & 66.39 & 23.54 & 55.25 & 50.39 \\
\midrule
GPT-3.5-turbo-16k (full, 1-shot) & 51.19 & 72.89 & 62.62 & 60.49 \\
GPT-3.5-turbo-16k (section, 1-shot) & 51.22 & 44.59 & 72.44 & 70.03 \\
GPT-4 (full, 1-shot) & 70.72 & 72.98 & 71.88 & 67.08 \\
GPT-4 (section, 1-shot) & 70.57 & 60.35 & \bf 78.48 & \bf 71.51 \\
\midrule
GPT-3.5-turbo-16k (full, 2-shot) & 51.77 & \bf 77.97 & 63.08 & 61.74 \\
GPT-3.5-turbo-16k (section, 2-shot) & 49.73 & 61.51 & 72.42 & 70.18 \\
GPT-4-32k (full, 2-shot) & 60.90 & 74.35 & 71.09 & 65.39 \\
GPT-4-32k (section, 2-shot) & 71.22 & 69.21 & 70.81 & 67.28 \\
\bottomrule
\toprule
\multicolumn{5}{c}{\bf Objective-Results} \\
\midrule
\bf Model 
& \bf Claim Recall & \bf Claim Prec & \bf Citation Recall & \bf Citation Prec \\
\midrule
BART + SAMSum FT (full) & 3.12 & 25.00 & -- & -- \\ 
BART + SAMSum FT (section) & 33.96 & 11.44 & -- & -- \\ 
BioBART (full) & 0.00 & 0.00 & -- & -- \\ 
BioBART (section) & 34.37 & 15.47 & -- & -- \\
\midrule
GPT-3.5-turbo (full, 0-shot) & 64.24 & 48.21 & 49.33 & 47.74 \\
GPT-3.5-turbo (section, 0-shot) & 66.25 & 5.98 & 56.11 & 44.93 \\
GPT-4 (full, 0-shot) & 71.60 & 57.86 & 62.08 & 60.62 \\
GPT-4 (section, 0-shot) & 76.67 & 8.60 & 65.84 & 46.66 \\
\midrule
GPT-3.5-turbo-16k (full, 1-shot) & 61.83 & 57.91 & 77.57 & 77.48 \\
GPT-3.5-turbo-16k (section, 1-shot) & 64.00 & 13.80 & 76.93 & 69.80 \\
GPT-4 (full, 1-shot) & 71.66 & \bf 75.44 & \bf 80.88 & \bf 78.68 \\
GPT-4 (section, 1-shot) & 77.66 & 25.94 & 77.27 & 73.25 \\
\midrule
GPT-3.5-turbo-16k (full, 2-shot) & 61.83 & 62.83 & 79.46 & 79.46 \\
GPT-3.5-turbo-16k (section, 2-shot) & 70.66 & 24.57 & 69.94 & 65.72 \\
GPT-4-32k (full, 2-shot) & 59.39 & 75.21 & 77.59 & 74.71 \\
GPT-4-32k (section, 2-shot) & 80.16 & 22.23 & 67.71 & 64.97 \\
\bottomrule
\toprule
\multicolumn{5}{c}{\bf Assessment and Plan} \\
\midrule
\bf Model 
& \bf Claim Recall & \bf Claim Prec & \bf Citation Recall & \bf Citation Prec \\
\midrule
BART + SAMSum FT (full) & 0.63 & 10.71 & -- & -- \\ 
BART + SAMSum FT (section) & 20.01 & 27.03 & -- & -- \\ 
BioBART (full) & 0.00 & 0.00 & -- & -- \\ 
BioBART (section) & 17.84 & 24.36 & -- & -- \\
\midrule
GPT-3.5-turbo (full, 0-shot) & 52.47 & 73.10 & 55.53 & 41.54 \\
GPT-3.5-turbo (section, 0-shot) & 60.32 & 53.43 & 59.98 & 46.24 \\
GPT-4 (full, 0-shot) & 53.90 & 82.74 & \bf 67.77 & \bf 65.43 \\
GPT-4 (section, 0-shot) & 66.60 & 50.73 & 57.81 & 51.22 \\
\midrule
GPT-3.5-turbo-16k (full, 1-shot) & 58.40 & \bf 74.43 & 62.58 & 58.29 \\
GPT-3.5-turbo-16k (section, 1-shot) & 63.90 & 46.80 & 62.63 & 57.43 \\
GPT-4 (full, 1-shot) & 61.07 & 69.59 & 62.83 & 55.12 \\
GPT-4 (section, 1-shot) & 67.38 & 50.71 & 63.04 & 53.40 \\
\midrule
GPT-3.5-turbo-16k (full, 2-shot) & 62.53 & 66.48 & 57.48 & 52.79 \\
GPT-3.5-turbo-16k (section, 2-shot) & 66.58 & 50.09 & 62.31 & 58.23 \\
GPT-4-32k (full, 2-shot) & 65.87 & 62.90 & 60.29 & 49.26 \\
GPT-4-32k (section, 2-shot) & \bf 72.11 & 50.99 & 63.28 & 54.44 \\
\bottomrule
\end{tabular}
}
\caption{Note generation results of each section on ACI-BENCH-test1, evaluated with \modelname (GPT-4).}
\label{tab:ACI_section}
\end{table*}

\begin{table*}[!ht]
\centering
\resizebox{0.55\linewidth}{!}{
\begin{tabular}{lcccc}
\toprule
\multicolumn{5}{l}{\textbf{Evaluation Method:} \modelname computed with Mistral} \\
\bottomrule
\toprule
\multicolumn{5}{c}{\bf Subjective} \\
\midrule
\bf Model & \bf Claim Recall & \bf Claim Prec & \bf Citation Recall & \bf Citation Prec \\
\midrule
BART + SAMSum FT (full) & 55.52 & 53.92 & -- & -- \\
BART + SAMSum FT (section) & 51.05 & 58.00 & -- & -- \\
BioBART (full) & 44.33 & 48.37 & -- & -- \\
BioBART (section) & 40.21 & 57.67 & -- & -- \\
\midrule
GPT-3.5-turbo (full, 0-shot) & 41.43 & 75.12 & 95.30 & 76.89 \\
GPT-3.5-turbo (section, 0-shot) & 57.64 & 45.27 & 86.02 & 61.57 \\
GPT-4 (full, 0-shot) & 41.36 & \bf 78.44 & \bf 97.92 & \bf 90.71 \\
GPT-4 (section, 0-shot) & \bf 65.99 & 69.35 & 93.05 & 85.15 \\
\midrule
GPT-3.5-turbo-16k (full, 1-shot) & 49.52 & 74.05 & 94.60 & 88.57 \\
GPT-3.5-turbo-16k (section, 1-shot) & 59.74 & 51.08 & 85.45 & 79.70 \\
GPT-4 (full, 1-shot) & 58.96 & 76.46 & 97.82 & 86.50 \\
GPT-4 (section, 1-shot) & 63.35 & 65.57 & 95.42 & 83.76 \\
\midrule
GPT-3.5-turbo-16k (full, 2-shot) & 61.44 & 72.30 & 93.14 & 82.31 \\
GPT-3.5-turbo-16k (section, 2-shot) & 62.42 & 49.08 & 89.06 & 80.16 \\
GPT-4-32k (full, 2-shot) & 59.61 & 73.09 & 96.11 & 83.89 \\
GPT-4-32k (section, 2-shot) & 65.39 & 67.61 & 95.12 & 83.54 \\
\bottomrule
\toprule
\multicolumn{5}{c}{\bf Objective-Exam} \\
\midrule
\bf Model & \bf Claim Recall & \bf Claim Prec & \bf Citation Recall & \bf Citation Prec \\
\midrule
BART + SAMSum FT (full) & 5.18 & 42.41 & -- & -- \\
BART + SAMSum FT (section) & 45.81 & 31.56 & -- & -- \\
BioBART (full) & 2.50 & 30.00 & -- & -- \\
BioBART (section) & 44.91 & 24.69 & -- & -- \\
\midrule
GPT-3.5-turbo (full, 0-shot) & 73.15 & 78.74 & 89.92 & 78.55 \\
GPT-3.5-turbo (section, 0-shot) & 71.14 & 34.50 & 82.07 & 68.32 \\
GPT-4 (full, 0-shot) & 74.20 & 74.68 & 90.79 & 87.10 \\
GPT-4 (section, 0-shot) & 76.97 & 30.21 & 84.59 & 76.08 \\
\midrule
GPT-3.5-turbo-16k (full, 1-shot) & 64.38 & 78.41 & 82.09 & 79.79 \\
GPT-3.5-turbo-16k (section, 1-shot) & 61.78 & 48.42 & 89.92 & 83.38 \\
GPT-4 (full, 1-shot) & 80.37 & \bf 80.74 & 84.08 & 80.37 \\
GPT-4 (section, 1-shot) & 78.66 & 64.75 & 91.50 & 86.53 \\
\midrule
GPT-3.5-turbo-16k (full, 2-shot) & 71.24 & 82.35 & 79.46 & 75.82 \\
GPT-3.5-turbo-16k (section, 2-shot) & 62.78 & 65.96 & \bf 91.62 & \bf 89.52 \\
GPT-4-32k (full, 2-shot) & 70.77 & 72.77 & 87.52 & 83.83 \\
GPT-4-32k (section, 2-shot) & \bf 82.69 & 70.04 & 85.33 & 81.19 \\
\bottomrule
\toprule
\multicolumn{5}{c}{\bf Objective-Results} \\
\midrule
\bf Model & \bf Claim Recall & \bf Claim Prec & \bf Citation Recall & \bf Citation Prec \\
\midrule
BART + SAMSum FT (full) & 3.12 & 25.00 & -- & -- \\
BART + SAMSum FT (section) & 38.19 & 16.53 & -- & -- \\
BioBART (full) & 0.00 & 0.00 & -- & -- \\
BioBART (section) & 40.93 & 18.60 & -- & -- \\
\midrule
GPT-3.5-turbo (full, 0-shot) & 81.53 & 62.38 & 91.03 & 85.56 \\
GPT-3.5-turbo (section, 0-shot) & 70.68 & 10.11 & 81.46 & 64.97 \\
GPT-4 (full, 0-shot) & 89.31 & 68.07 & \bf 95.00 & \bf 93.54 \\
GPT-4 (section, 0-shot) & 85.26 & 13.67 & 93.57 & 86.36 \\
\midrule
GPT-3.5-turbo-16k (full, 1-shot) & 79.27 & 67.52 & 82.61 & 82.21 \\
GPT-3.5-turbo-16k (section, 1-shot) & 77.20 & 20.14 & 92.04 & 77.56 \\
GPT-4 (full, 1-shot) & 86.19 & \bf 85.29 & 85.29 & 85.29 \\
GPT-4 (section, 1-shot) & \bf 90.78 & 31.37 & 89.74 & 83.96 \\
\midrule
GPT-3.5-turbo-16k (full, 2-shot) & 77.45 & 69.17 & 81.22 & 81.22 \\
GPT-3.5-turbo-16k (section, 2-shot) & 79.62 & 31.93 & 86.67 & 83.26 \\
GPT-4-32k (full, 2-shot) & 76.84 & 84.90 & 79.31 & 77.59 \\
GPT-4-32k (section, 2-shot) & 89.32 & 27.42 & 78.03 & 76.12 \\
\bottomrule
\toprule
\multicolumn{5}{c}{\bf Assessment and Plan} \\
\midrule
\bf Model & \bf Claim Recall & \bf Claim Prec & \bf Citation Recall & \bf Citation Prec \\
\midrule
BART + SAMSum FT (full) & 1.63 & 23.21 & -- & -- \\
BART + SAMSum FT (section) & 30.72 & 33.83 & -- & -- \\
BioBART (full) & 0.00 & 0.00 & -- & -- \\
BioBART (section) & 26.23 & 31.96 & -- & -- \\
\midrule
GPT-3.5-turbo (full, 0-shot) & 72.63 & 80.30 & 88.99 & 79.27 \\
GPT-3.5-turbo (section, 0-shot) & 76.39 & 60.97 & 83.82 & 61.93 \\
GPT-4 (full, 0-shot) & 74.00 & \bf 84.77 & \bf 92.30 & \bf 87.23 \\
GPT-4 (section, 0-shot) & 77.24 & 58.16 & 86.66 & 74.76 \\
\midrule
GPT-3.5-turbo-16k (full, 1-shot) & 71.30 & 77.53 & 83.17 & 80.53 \\
GPT-3.5-turbo-16k (section, 1-shot) & 71.82 & 58.50 & 87.16 & 80.12 \\
GPT-4 (full, 1-shot) & 74.05 & 78.01 & 88.29 & 79.82 \\
GPT-4 (section, 1-shot) & 76.12 & 59.77 & 88.70 & 74.45 \\
\midrule
GPT-3.5-turbo-16k (full, 2-shot) & 70.17 & 73.26 & 84.16 & 77.29 \\
GPT-3.5-turbo-16k (section, 2-shot) & 69.48 & 60.30 & 85.91 & 74.66 \\
GPT-4-32k (full, 2-shot) & 74.89 & 70.89 & 86.16 & 74.26 \\
GPT-4-32k (section, 2-shot) & \bf 77.43 & 60.72 & 86.11 & 69.43 \\
\bottomrule
\end{tabular}
}
\caption{Note generation results on different sections of ACI-BENCH-test1 evaluated with \modelname (Mistral).}
\label{tab:ACI_Mistral}
\end{table*}

\begin{table*}[!ht]
\centering
\resizebox{0.55\linewidth}{!}{
\begin{tabular}{lcccc}
\toprule
\multicolumn{5}{l}{\textbf{Evaluation Method:} \modelname computed with TRUE} \\
\bottomrule
\toprule
\multicolumn{5}{c}{\bf Subjective} \\
\midrule
\bf Model & \bf Claim Recall & \bf Claim Prec & \bf Citation Recall & \bf Citation Prec \\
\midrule
BART + SAMSum FT (full) & 45.74 & 49.26 & -- & -- \\
BART + SAMSum FT (section) & 40.99 & 52.31 & -- & -- \\
BioBART (full) & 37.63 & 42.04 & -- & -- \\
BioBART (section) & 32.20 & 47.37 & -- & -- \\
\midrule
GPT-3.5-turbo (full, 0-shot) & 28.40 & \bf 67.75 & 61.31 & 40.40 \\
GPT-3.5-turbo (section, 0-shot) & 45.33 & 33.87 & 54.97 & 30.80 \\
GPT-4 (full, 0-shot) & 31.09 & 72.18 & \bf 66.95 & \bf 45.98 \\
GPT-4 (section, 0-shot) & 50.03 & 64.72 & 51.55 & 41.39 \\
\midrule
GPT-3.5-turbo-16k (full, 1-shot) & 41.29 & 65.96 & 44.94 & 38.99 \\
GPT-3.5-turbo-16k (section, 1-shot) & 56.08 & 43.66 & 44.56 & 40.29 \\
GPT-4 (full, 1-shot) & 47.75 & 69.38 & 52.98 & 40.39 \\
GPT-4 (section, 1-shot) & \bf 64.72 & 60.22 & 51.88 & 40.42 \\
\midrule
GPT-3.5-turbo-16k (full, 2-shot) & 50.12 & 63.41 & 43.08 & 37.14 \\
GPT-3.5-turbo-16k (section, 2-shot) & 43.61 & 40.38 & 48.94 & 40.33 \\
GPT-4-32k (full, 2-shot) & 53.54 & 64.99 & 46.25 & 37.32 \\
GPT-4-32k (section, 2-shot) & 63.42 & 62.43 & 44.82 & 35.93 \\
\bottomrule
\toprule
\multicolumn{5}{c}{\bf Objective-Exam} \\
\midrule
\bf Model & \bf Claim Recall & \bf Claim Prec & \bf Citation Recall & \bf Citation Prec \\
\midrule
BART + SAMSum FT (full) & 1.71 & 24.17 & -- & -- \\
BART + SAMSum FT (section) & 34.17 & 26.14 & -- & -- \\
BioBART (full) & 1.46 & 15.83 & -- & -- \\
BioBART (section) & 34.96 & 18.28 & -- & -- \\
\midrule
GPT-3.5-turbo (full, 0-shot) & 61.82 & 62.06 & 69.54 & 55.63 \\
GPT-3.5-turbo (section, 0-shot) & 67.47 & 23.90 & 58.99 & 41.60 \\
GPT-4 (full, 0-shot) & 67.27 & 71.48 & 71.75 & 63.96 \\
GPT-4 (section, 0-shot) & 71.44 & 21.30 & 45.65 & 35.35 \\
\midrule
GPT-3.5-turbo-16k (full, 1-shot) & 49.95 & 72.29 & 58.90 & 54.13 \\
GPT-3.5-turbo-16k (section, 1-shot) & 52.49 & 44.54 & 68.66 & 61.92 \\
GPT-4 (full, 1-shot) & 74.53 & 68.91 & 66.58 & 59.37 \\
GPT-4 (section, 1-shot) & 71.91 & 60.81 & 71.67 & 63.13 \\
\midrule
GPT-3.5-turbo-16k (full, 2-shot) & 55.00 & \bf 79.64 & 60.08 & 53.18 \\
GPT-3.5-turbo-16k (section, 2-shot) & 52.84 & 59.67 & \bf 72.72 & \bf 69.34 \\
GPT-4-32k (full, 2-shot) & 65.50 & 72.58 & 62.30 & 57.88 \\
GPT-4-32k (section, 2-shot) & \bf 75.61 & 66.35 & 63.35 & 57.35 \\
\bottomrule
\toprule
\multicolumn{5}{c}{\bf Objective-Results} \\
\midrule
\bf Model & \bf Claim Recall & \bf Claim Prec & \bf Citation Recall & \bf Citation Prec \\
\midrule
BART + SAMSum FT (full) & 3.12 & 25.00 & -- & -- \\
BART + SAMSum FT (section) & 35.89 & 12.81 & -- & -- \\
BioBART (full) & 0.00 & 0.00 & -- & -- \\
BioBART (section) & 35.90 & 15.71 & -- & -- \\
\midrule
GPT-3.5-turbo (full, 0-shot) & 56.50 & 51.92 & 74.36 & 70.67 \\
GPT-3.5-turbo (section, 0-shot) & 64.42 & 4.86 & 54.98 & 34.51 \\
GPT-4 (full, 0-shot) & 63.87 & 60.15 & \bf 85.83 & \bf 80.21 \\
GPT-4 (section, 0-shot) & \bf 81.64 & 8.73 & 55.91 & 44.80 \\
\midrule
GPT-3.5-turbo-16k (full, 1-shot) & 51.31 & 61.32 & 72.97 & 72.07 \\
GPT-3.5-turbo-16k (section, 1-shot) & 66.27 & 14.49 & 73.43 & 53.60 \\
GPT-4 (full, 1-shot) & 65.17 & \bf 78.24 & 72.06 & 69.85 \\
GPT-4 (section, 1-shot) & 78.07 & 27.31 & 70.31 & 62.20 \\
\midrule
GPT-3.5-turbo-16k (full, 2-shot) & 52.34 & 65.83 & 71.36 & 71.36 \\
GPT-3.5-turbo-16k (section, 2-shot) & 71.36 & 28.26 & 66.54 & 62.16 \\
GPT-4-32k (full, 2-shot) & 52.95 & 77.40 & 62.07 & 59.77 \\
GPT-4-32k (section, 2-shot) & 81.51 & 22.96 & 64.12 & 61.07 \\
\bottomrule
\toprule
\multicolumn{5}{c}{\bf Assessment and Plan} \\
\midrule
\bf Model & \bf Claim Recall & \bf Claim Prec & \bf Citation Recall & \bf Citation Prec \\
\midrule
BART + SAMSum FT (full) & 0.42 & 10.71 & -- & -- \\
BART + SAMSum FT (section) & 17.79 & 23.11 & -- & -- \\
BioBART (full) & 0.00 & 0.00 & -- & -- \\
BioBART (section) & 15.47 & 19.78 & -- & -- \\
\midrule
GPT-3.5-turbo (full, 0-shot) & 44.92 & 63.23 & \bf 68.00 & 55.74 \\
GPT-3.5-turbo (section, 0-shot) & 55.00 & 49.47 & 55.87 & 31.16 \\
GPT-4 (full, 0-shot) & 47.68 & \bf 74.80 & 65.12 & \bf 57.98 \\
GPT-4 (section, 0-shot) & 57.29 & 45.77 & 44.39 & 32.83 \\
\midrule
GPT-3.5-turbo-16k (full, 1-shot) & 53.91 & 63.62 & 50.45 & 45.21 \\
GPT-3.5-turbo-16k (section, 1-shot) & 58.22 & 42.88 & 47.14 & 36.30 \\
GPT-4 (full, 1-shot) & 56.30 & 63.99 & 54.01 & 40.78 \\
GPT-4 (section, 1-shot) & 64.30 & 46.09 & 50.13 & 30.77 \\
\midrule
GPT-3.5-turbo-16k (full, 2-shot) & 56.80 & 58.82 & 47.48 & 39.05 \\
GPT-3.5-turbo-16k (section, 2-shot) & 61.51 & 44.62 & 50.14 & 36.89 \\
GPT-4-32k (full, 2-shot) & 60.10 & 55.73 & 49.11 & 34.36 \\
GPT-4-32k (section, 2-shot) & \bf 67.65 & 44.78 & 49.03 & 29.61 \\
\bottomrule
\end{tabular}
}
\caption{Note generation results on different sections of ACI-BENCH-test1 evaluated with \modelname (TRUE).}
\label{tab:ACI_TRUE}
\end{table*}

\begin{table*}[ht]
\centering
\resizebox{0.7\textwidth}{!}{
\begin{tabular}{lcccc}
\toprule
\multicolumn{5}{l}{\textbf{Evaluation Methods:} Rouge, BERTScore, BLEU, RadGraph} \\
\bottomrule
\toprule
\bf Model/Prompt Style  
& \bf Rouge-L & \bf BERTSore & \bf BLEU & \bf F1-RadGraph \\
\midrule
GPT-3.5-turbo (0-shot) & 28.83 & 86.33 & 8.22 & 25.93 \\
GPT-4 (0-shot) & 28.98 & 86.39 & 7.81 & 25.47 \\
\midrule
GPT-3.5-turbo (6-shot) & 34.27 & 87.46 & 13.10 & \bf 29.11 \\
GPT-4 (6-shot) & 34.25 & 87.42 & 12.59 & 28.13 \\
\midrule
GPT-3.5-turbo-16k (12-shot) & 34.07 & \bf 87.56 & \bf 13.99 & 28.69 \\
GPT-4 (12-shot) & \bf 34.61 & 87.54 & 13.12 & 28.51 \\
\bottomrule
\end{tabular}
}
\caption{Report summarization performance on 200 examples in MIMIC-III evaluated with existing metrics. The examples are proportionally sampled from each modality. We select one or two training example(s) from each of the 6 modality-anatomy pairs that contain a train set as the few-shot demos.}
\label{tab:MIMIC_existing}
\end{table*}

\begin{table*}[ht]
\centering
\resizebox{0.8\textwidth}{!}{
\begin{tabular}{lcccc}
\toprule
\multicolumn{5}{l}{\textbf{Evaluation Method:} \modelname computed with GPT-4} \\
\bottomrule
\toprule
\bf Model/Prompt Style  
& \bf Claim Recall & \bf Claim Prec & \bf Citation Recall & \bf Citation Prec \\
\midrule
GPT-3.5-turbo (0-shot) & 62.40 & 24.20 & 91.63 & 89.21 \\
GPT-4 (0-shot) & \bf 63.39 & 25.38 & 97.48 & 96.64 \\
\midrule
GPT-3.5-turbo (6-shot) & 53.19 & 27.82 & 98.14 & 96.57 \\
GPT-4 (6-shot)& 57.12 & 29.12 & \bf 99.11 & \bf 97.98 \\
\midrule
GPT-3.5-turbo-16k (12-shot) & 47.38 & 29.77 & 97.37 & 96.26 \\
GPT-4 (12-shot) & 56.01 & \bf 29.98 & 97.73 & 96.13 \\
\bottomrule
\end{tabular}
}
\caption{Report summarization results on 200 examples in MIMIC-III evaluated with \modelname (GPT-4).}
\label{tab:MIMIC_orig}
\end{table*}

\begin{table*}[!ht]
    \centering
    \resizebox{0.8\linewidth}{!}{
    \begin{tabular}{lcccc}
    \toprule
    \multicolumn{5}{l}{\textbf{Evaluation Method:} \modelname computed with Mistral} \\
    \bottomrule
    \toprule
    \bf Model & \bf Claim Recall & \bf Claim Prec & \bf Citation Recall & \bf Citation Prec \\
    \midrule
    GPT-3.5-turbo (0-shot) & 67.36 & 38.78 & 98.17 & 86.71 \\
    GPT-4 (0-shot) & \bf 69.26 & 39.47 & 99.79 & \bf 97.91 \\
    \midrule
    GPT-3.5-turbo (6-shot) & 59.60 & 43.09 & 99.50 & 96.26 \\
    GPT-4 (6-shot) & 64.99 & 43.99 & \bf 99.93 & 97.24 \\
    \midrule
    GPT-3.5-turbo-16k (12-shot) & 59.79 & 44.28 & 99.59 & 95.94 \\
    GPT-4 (12-shot) & 62.64 & \bf 45.72 & 99.43 & 96.51 \\
    \bottomrule
    \end{tabular}
    }
    \caption{Report summarization performance on 200 examples in MIMIC-III evaluated with \modelname (Mistral).}
    \label{tab:MIMIC_Mistral}
    \end{table*}

\begin{table*}[!ht]
\centering
\resizebox{0.8\linewidth}{!}{
\begin{tabular}{lcccc}
\toprule
\multicolumn{5}{l}{\textbf{Evaluation Method:} \modelname computed with TRUE} \\
\bottomrule
\toprule
\bf Model & \bf Claim Recall & \bf Claim Prec & \bf Citation Recall & \bf Citation Prec \\
\midrule
GPT-3.5-turbo (0-shot) & \bf 47.55 & 17.95 & 96.68 & 64.24 \\
GPT-4 (0-shot) & 46.00 & 17.96 & 97.54 & \bf 93.86 \\
\midrule
GPT-3.5-turbo (6-shot) & 38.60 & 21.49 & 96.67 & 92.43 \\
GPT-4 (6-shot) & 42.47 & 22.05 & \bf 97.69 & 91.81 \\
\midrule
GPT-3.5-turbo-16k (12-shot) & 35.12 & 23.54 & 95.42 & 89.00 \\
GPT-4 (12-shot) & 43.17 & \bf 23.70 & 95.80 & 89.87 \\
\bottomrule
\end{tabular}
}
\caption{Report summarization performance on 200 examples in MIMIC-III evaluated with \modelname (TRUE).}
\label{tab:MIMIC_TRUE}
\end{table*}

\begin{table*}[ht]
\centering
\resizebox{0.7\textwidth}{!}{
\begin{tabular}{lcccc}
\toprule
\multicolumn{5}{l}{\textbf{Evaluation Methods:} Rouge, BERTScore, BLEU, MEDCON} \\
\bottomrule
\toprule
\bf Model/Prompt Style  
& \bf Rouge-L & \bf BERTSore & \bf BLEU & \bf MEDCON \\
\midrule
GPT-3.5-turbo (0-shot) & 28.26 & 91.20 & 8.36 & 44.93 \\
GPT-4 (0-shot) & 31.02 & 91.41 & 6.69 & \bf 49.08 \\
\midrule
GPT-3.5-turbo (1-shot) & 29.80 & 91.56 & \bf 8.74 & 44.34 \\
GPT-4 (1-shot) & \bf 33.15 & \bf 91.85 & 8.36 & 46.14 \\
\midrule
GPT-3.5-turbo-16k (2-shot) & 29.98 & 91.34 & 8.57 & 44.64 \\
GPT-4 (2-shot) & 32.96 & 91.78 & 8.50 & 48.42 \\
\bottomrule
\end{tabular}
}
\caption{Question summarization performance on MeQSum evaluated with existing metrics. We experiment on the test set provided by the MEDIQA 2021 challenge~\cite{ben-abacha-etal-2021-overview} with 100 examples.}
\label{tab:meqsum_existing}
\end{table*}

\begin{table*}[ht]
\centering
\resizebox{0.8\textwidth}{!}{
\begin{tabular}{lcccc}
\toprule
\multicolumn{5}{l}{\textbf{Evaluation Method:} \modelname computed with GPT-4} \\
\bottomrule
\toprule
\bf Model/Prompt Style  
& \bf Claim Recall & \bf Claim Prec & \bf Citation Recall & \bf Citation Prec \\
\midrule
GPT-3.5-turbo (0-shot) & 49.00 & 48.00 & 85.00 & 77.93 \\
GPT-4 (0-shot) & 52.00 & 53.00 & 93.00 & \bf 86.90 \\
\midrule
GPT-3.5-turbo (1-shot) & 44.00 & 48.00 & 87.00 & 81.72 \\
GPT-4 (1-shot) & 50.00 & \bf 56.00 & \bf 94.00 & 85.27 \\
\midrule
GPT-3.5-turbo-16k (2-shot) & 46.00 & 47.00 & 85.00 & 80.72 \\
GPT-4 (2-shot) & \bf 53.00 & 49.00 & 93.00 & 86.42 \\
\bottomrule
\end{tabular}
}
\caption{Question summarization performance on MeQSum evaluated with \modelname computed with GPT-4.}
\label{tab:meqsum}
\end{table*}

\begin{table*}[!ht]
    \centering
    \resizebox{0.8\linewidth}{!}{
    \begin{tabular}{lcccc}
    \toprule
    \multicolumn{5}{l}{\textbf{Evaluation Method:} \modelname computed with Mistral} \\
    \bottomrule
    \toprule
    \bf Model & \bf Claim Recall & \bf Claim Prec & \bf Citation Recall & \bf Citation Prec \\
    \midrule
    GPT-3.5-turbo (0-shot) & 64.00 & 64.00 & 84.00 & 69.07 \\
    GPT-4 (0-shot) & 71.00 & 72.00 & 86.00 & 69.20 \\
    \midrule
    GPT-3.5-turbo (1-shot) & 71.00 & 64.00 & \bf 88.00 & 66.77 \\
    GPT-4 (6-shot) & \bf 79.00 & 76.00 & \bf 88.00 & \bf 69.58 \\
    \midrule
    GPT-3.5-turbo-16k (2-shot) & 74.00 & \bf 79.00 & 82.00 & 72.37 \\
    GPT-4 (12-shot) & 72.00 & 73.00 & 87.00 & \bf 69.58 \\
    \bottomrule
    \end{tabular}
    }
    \caption{Question summarization performance on MeQSum (MEDIQA 2021 test set) evaluated with \modelname (Mistral).}
    \label{tab:meqsum_Mistral}
    \end{table*}

\begin{table*}[!ht]
\centering
\resizebox{0.8\linewidth}{!}{
\begin{tabular}{lcccc}
\toprule
\multicolumn{5}{l}{\textbf{Evaluation Method:} \modelname computed with TRUE} \\
\bottomrule
\toprule
\bf Model & \bf Claim Recall & \bf Claim Prec & \bf Citation Recall & \bf Citation Prec \\
\midrule
GPT-3.5-turbo (0-shot) & 29.00 & 14.00 & 69.00 & 57.23 \\
GPT-4 (0-shot) & \bf 39.00 & 10.00 & 82.00 & 66.22 \\
\midrule
GPT-3.5-turbo (1-shot) & 30.00 & 12.00 & 68.00 & 55.49 \\
GPT-4 (6-shot) & 33.00 & \bf 16.00 & \bf 86.00 & \bf 67.08 \\
\midrule
GPT-3.5-turbo-16k (2-shot) & 33.00 & 14.00 & 69.00 & 58.19 \\
GPT-4 (12-shot) & 37.00 & 11.00 & 80.00 & 66.92 \\
\bottomrule
\end{tabular}
}
\caption{Question summarization performance on MeQSum (MEDIQA 2021 test set) evaluated with \modelname (TRUE).}
\label{tab:meqsum_TRUE}
\end{table*}

\begin{table*}[ht]
\centering
\resizebox{0.8\linewidth}{!}{
\begin{tabular}{lccc|c}
\toprule
\bf Model/Prompt Style  
& \bf Rouge-L & \bf BLEU & \bf F1-RadGraph & \bf Claim Recall (GPT-4) \\
\midrule
GPT-4 (no citation, 6-shot) & 33.95 & \underline{13.01} & \underline{29.08} & 55.24  \\
GPT-4 (with citation, 6-shot) & \underline{34.25} & 12.59 & 28.13 & \underline{57.12}  \\
\midrule
GPT-4 (no citation, 12-shot) & \underline{34.81} & \underline{14.40} & \underline{30.61} & 55.43  \\
GPT-4 (with citation, 12-shot) & 34.54 & 13.12 & 28.51 & \underline{56.01} \\
\bottomrule
\end{tabular}
}
\caption{Report summarization results with or without asking the model to generate citations. We report the results on 200 examples in MIMIC-III evaluated with \modelname computed with GPT-4. The examples are proportionally sampled from each modality. We underline the better result in each block.}
\label{tab:MIMIC_citation_or_not}
\end{table*}

\end{document}